\documentclass[acmtog]{acmart}

\usepackage{booktabs} 

\setcitestyle{square}

\usepackage[ruled]{algorithm2e} 

\SetAlFnt{\small}
\SetAlCapFnt{\small}
\SetAlCapNameFnt{\small}
\SetAlCapHSkip{0pt}
\IncMargin{-\parindent}

\acmJournal{TOG}

\setcopyright{none}

\newcommand{\etal}{et al.}
\usepackage{color}
\usepackage{soul}
\usepackage{breqn}
\usepackage{booktabs}
\usepackage{rotating}

\definecolor{purple}{rgb}{0.65,0,0.65}
\definecolor{blue}{rgb}{0, 0.2, 0.8}
\definecolor{orange}{rgb}{0.6, 0.6, 0}
\definecolor{red}{rgb}{0.8, 0.2, 0.2}
\definecolor{magenta}{rgb}{0.5, 0.0, 1.0}
\definecolor{black}{rgb}{0.0, 0.0, 0.0}
\definecolor{cyan}{rgb}{0, 0.65, 0.65}
\definecolor{olive}{rgb}{0.2, 0.6, 0.5}
\usepackage{xcolor}

\newif\ifdraft
\draftfalse

\ifdraft
\newcommand{\rzc}[1]{{\color{blue}\textbf{RZ:} #1}}
\newcommand{\dlc}[1]{{\color{magenta}\textbf{DL:} #1}}
\newcommand{\dcc}[1]{{\color{red}\textbf{DC:} #1}}
\newcommand{\kac}[1]{{\color{orange}\textbf{KA:} #1}}
\newcommand{\msc}[1]{{\color{green}\textbf{MS:} #1}}
\newcommand{\AAAc}[1]{{\color{blue}\textbf{AA:} #1}}
\newcommand{\tkc}[1]{{\color{green}\textbf{TK:} #1}}
\newcommand{\myc}[1]{{\color{cyan}\textbf{MY:} #1}}

\newcommand{\dl}[1]{{\color{magenta}#1}}

\else
\newcommand{\rzc}[1]{}
\newcommand{\dlc}[1]{}
\newcommand{\dcc}[1]{}
\newcommand{\kac}[1]{}
\newcommand{\msc}[1]{}
\newcommand{\AAAc}[1]{}
\newcommand{\tkc}[1]{}
\newcommand{\myc}[1]{}

\newcommand{\dl}[1]{{\color{black}#1}}

\fi

\newcommand{\bbp}{{\bf P}}
\newcommand{\bc}{{\bf c}}
\newcommand{\bff}{{\bf f}}

\newcommand{\bq}{{\bf q}}
\newcommand{\br}{{\bf r}}
\newcommand{\bs}{{\bf s}}

\newcommand{\Loss}{\mathcal{L}}

\newcommand{\cc}{\mathcal{C}}
\newcommand{\pp}{\mathcal{P}}
\newcommand{\qq}{\mathcal{Q}}
\newcommand{\bbe}{\mathbb{E}}
\def \figures {./figures/}

\begin{document}

\title{MotioNet: 3D Human Motion Reconstruction from Monocular Video with Skeleton Consistency}

\author{Mingyi Shi}
\affiliation{\institution{Shandong University}}
\affiliation{\institution{AICFVE, Beijing Film Academy}}  

\author{Kfir Aberman}
\affiliation{\institution{AICFVE, Beijing Film Academy}}
\affiliation{\institution{Tel-Aviv University}}

\author{Andreas Aristidou}
\orcid{0000-0001-7754-0791}
\affiliation{
  \institution{University of Cyprus}
  \streetaddress{75, Kallipoleos}
  \city{Nicosia}
  \state{Cyprus}
  \postcode{1678}
}
\affiliation{%
  \institution{RISE Research Centre}
}
\email{a.aristidou@ieee.org}
  
\author{Taku Komura}
\affiliation{%
  \institution{Edinburgh University}
  }
  
  \author{Dani Lischinski}
  \affiliation{\institution{Shandong University}}
\affiliation{%
  \institution{The Hebrew University of Jerusalem}
}
\affiliation{\institution{AICFVE, Beijing Film Academy}}  
  
  \author{Daniel Cohen-Or}
\affiliation{%
  \institution{Tel-Aviv University}
}
\affiliation{\institution{AICFVE, Beijing Film Academy}}  

  \author{Baoquan Chen}
\affiliation{ \institution{CFCS, Peking University}}
\affiliation{ \institution{AICFVE, Beijing Film Academy}}

\renewcommand\shortauthors{Shi, M. et al}

\begin{abstract}
We introduce \emph{MotioNet}, a deep neural network that directly reconstructs the motion of a 3D human skeleton from monocular video.
While previous methods \dl{rely on either rigging or} inverse kinematics (IK) to associate a consistent skeleton with temporally coherent joint rotations, our method is the first \dl{data-driven approach that directly outputs a kinematic skeleton, which is a complete, commonly used, motion representation}. 
At the crux of our approach lies a deep neural network with embedded kinematic priors, which decomposes sequences of 2D joint positions into two separate attributes: a single, symmetric, skeleton, encoded by bone lengths, and a sequence of 3D joint rotations associated with global root positions and foot contact labels. These attributes are fed into an integrated forward kinematics (FK) layer that outputs 3D positions, which are compared to a ground truth. In addition, an adversarial loss is applied to the velocities of the recovered rotations, to ensure that they lie on the manifold of natural joint rotations.
The key advantage of our approach is that it learns to infer natural joint rotations directly from the training data, rather than assuming an underlying model, or inferring them from joint positions using a data-agnostic IK solver.
We show that enforcing a single consistent skeleton along with temporally coherent joint rotations constrains the solution space, leading to a more robust handling of self-occlusions and depth ambiguities.

\end{abstract}

%

%
%



\begin{teaserfigure}
\centering
\includegraphics[width=\linewidth]{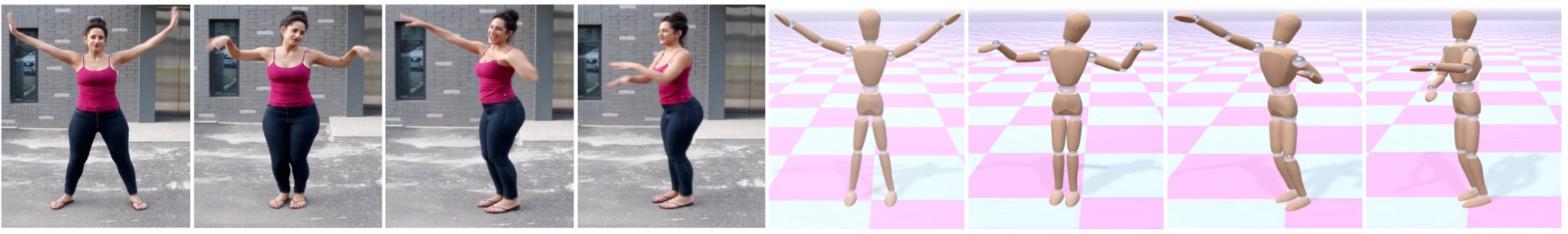} 
\caption{Given a monocular video of a performer, our approach, MotioNet, reconstructs a complete representation of the motion, consisting of a single symmetric skeleton, and a sequence of global root positions and 3D joint rotations. Thus, inverse kinematics is effectively integrated within the network, and is data-driven, rather than based on a universal prior. The images on the right were rendered from the output of our system after a simple rigging process.
} 
\label{fig:teaser}
\end{teaserfigure}

\maketitle

\section{Introduction}
\label{sec:intro}
Capturing the motion of humans has long been a fundamental task with a wide spectrum of applications in data-driven computer animation, special effects, gaming, activity recognition, and behavioral analysis. Motion is most accurately captured in a controlled setting using specialized hardware, such as magnetic trackers, depth sensors, or multi-camera optical systems. An alternative approach that has been researched extensively in recent years is to perform pose estimation and 3D motion reconstruction from ordinary monocular RGB video.

Motion capture from monocular video offers many advantages, such as a simple uncontrolled setup, low cost, and a non-intrusive capture process. While 3D human pose estimation is highly challenging due to depth ambiguities and occlusions, significant progress has been achieved in recent years by data-driven learning-based approaches. These approaches utilize deep neural networks to learn strong priors about the expected motion, which can significantly help with disambiguation and completion of missing data.

\begin{figure}
    \centering
    \includegraphics[width=1.0\columnwidth]{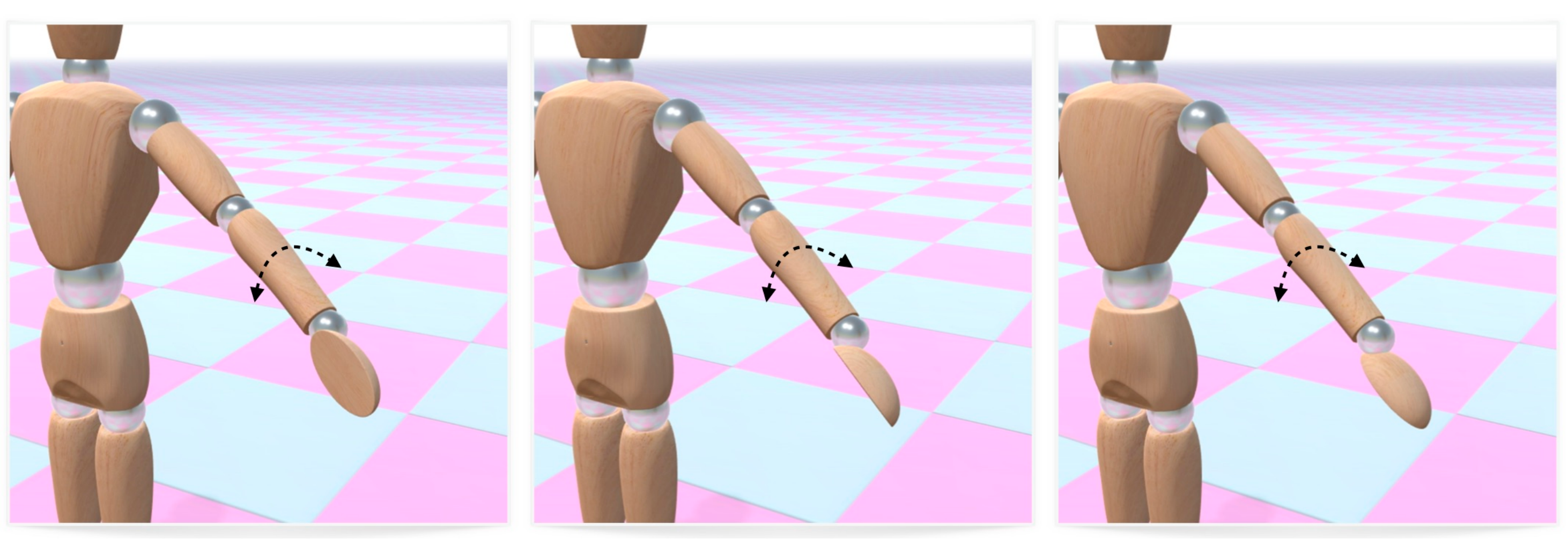}
    \caption{Joint rotation ambiguity. Given a set of fixed 3D joint positions, multiple limb rotations can connect every pair of consecutive joints. Thus, recovered 3D joint positions alone are not sufficient for driving a rigged and skinned virtual 3D character.
    }
    \label{fig:position_ambiguity}
\end{figure}

Given a video recording of a human motion, our ultimate goal is to reconstruct the motion in 3D space. 
\dl{One family of existing methods} extract a sequence of 3D poses from the video, where each pose is specified by the 3D location of each joint.
However, while the resulting representation may suffice for some applications, it is \emph{incomplete}. In particular, it does not contain all the information necessary to drive a rigged and skinned virtual 3D character, and the temporal consistency of the skeleton's bone lengths is not guaranteed.
While joint rotations may be recovered from joint positions via inverse kinematics (IK), the solution is generally not unique, as demonstrated in Figure~\ref{fig:position_ambiguity}. Furthermore, enforcing soft temporal coherence constraints over per-frame pose estimations may not ensure that the skeleton geometry remains invariant across all frames, and might result in unnatural movements.

\dl{Another group of works is aiming at the recovery of a parametric model that depicts the geometry of the body including joint rotations (see Section~\ref{sec:RelatedWork}). However, further rigging is required in order to extract a kinematic skeleton from such a model.} 

In this paper, we introduce \emph{MotioNet}, a deep neural network, trained to reconstruct the motion of a single performer from an ordinary monocular video (Figure \ref{fig:teaser}).
Instead of inferring a sequence of 3D joint positions, our network learns to extract a sequence of \emph{3D joint rotations} applied to a \emph{single} 3D skeleton. Thus, IK is effectively integrated within the network, and, consequently, is data-driven (learned).
Enforcing both a single skeleton and temporally coherent joint rotations not only constrains the solution space, ensuring consistency, but also leads to a more robust handling of self-occlusions and depth ambiguities.

To train our network, we leverage existing datasets that contain accurately captured full 3D human motions. Sequences of 3D poses are projected into 2D, and the network learns to decompose the resulting 2D joint position sequences into two separate attributes: a single, symmetric, skeleton, encoded by bone lengths, which define a geometric invariant along the entire sequence, and a sequence of 3D joint rotations, which capture the dynamic aspect of the motion. The 3D skeleton and joint rotations are fed into an integrated forward kinematics (FK) layer which applies the rotations successively along the bone hierarchy to reconstruct the original 3D motion sequence. In addition to the above, our network predicts a sequence of global positions of the root joint, as well as foot contact labels, because of the perceptual importance of the latter.

The network loss is a combination of terms that account for the bone lengths of the 3D skeleton, root global positions and foot contact labels, as well as joint positions recovered by the FK layer. While these attributes are compared to the ground truth 3D motion, the joint rotations are learned using an adversarial loss, which encourages their velocities to have a distribution of natural rotations.
In addition, in order to mitigate foot skating artifacts, we add a foot contact loss, to encourage the velocity of each foot to be zero in frames where it should be in contact with the ground.

A key advantage of our approach is that it does not require an IK step, which is data-agnostic and assumes an underlying constrained model. Instead, the task is integrated into the network, which learns to infer joint rotations directly from  training data of real human motions, rather than solving for them. 
Furthermore, as our system represents motion in the space of temporal convolutional filters, the learned motions are naturally smooth. All of this leads to a more data-driven human motion reconstruction, via the FK layer.

In order to bridge the gap between the training data and videos in the wild, we inject joint positional noise to the training input sequences and augment them with confidence values whose distribution mimics that of confidence values extracted by \cite{Cao:2018} from a variety of real videos. This augmentation step constitutes a regularizer in the space of the solutions, improves the stability of the results, and increases the robustness to occlusions.
An extensive set of experiments and ablation studies that we conducted to study the performance of our system and its different components, demonstrate the  quality and stability of our end-to-end, fully data-driven approach for monocular motion extraction.

%

%

\section{Related Work}
\label{sec:RelatedWork}
Prior to the deep learning era, various works performed 3D human pose estimation \dl{by performing exemplar-based IK to a set of constraints such as 2D keypoints~\cite{Grochow:2004}, or by fitting a skeleton to a voxel-based 3D model that was extracted over time from a video, e.g.,~\cite{Vlasic:2008,Liu:2013}. 
The latter approaches are broadly divided into generative methods, which reconstruct human pose by fitting a template model to the observed data~\cite{Wei:2012,Ye:2014}; discriminative methods, which cluster pixels to hypothesize body joint positions, fit a model, and then track the skeleton~\cite{Shotton:2013,Sharp:2012}; and hybrid methods, which combine the aforementioned techniques to achieve higher accuracy~\cite{Baak:2011}. See Sarafianos~\etal~\shortcite{Sarafianos:2016} for a detailed overview of 3D pose estimation methods.}


\paragraph{\textbf{3D Joint Position Reconstruction with Deep Learning:}}

\dl{Recently, enormous amount of work has been devoted to the development of deep learning networks that extract the 3D joint positions of the human body from an image or a video sequence;} 
Some methods, which extract the poses directly from the image pixels, have been popular due to the availability of video datasets that are associated with ground truth motion capture data (e.g., Human3.6M \cite{Ionescu:2014} and HumanEva \cite{Sigal:2009}). These methods use supervised learning to estimate the 3D joint positions~\cite{Toshev:2014,Zhou:2016,Tekin:2016,Pavlakos:2017,Tome:2017}. However, these 3D human pose estimation methods do not employ \dl{rigid body constraints}. Thus, some works integrate an iterative error feedback to regress bones and improve 3D pose estimation~\cite{Carreira:2016,Liang:2018,Luvizon:2018}. The main issue with such methods is that training data is captured and annotated in a controlled environment, thus they do not generalize well to the real world. To address this issue, some works use weak supervision, where both motion capture data and images in the wild are used~\cite{Mehta:2017_3DV}. \dl{In addition, most of these} works operate on a per-frame basis, which might lead to temporal incoherence and oscillations in motion. There have been efforts in exploiting temporal information from video to produce more robust predictions~\cite{Tekin:2016,Katircioglu:2018}, e.g., using long short-term memory (LSTM) networks \cite{Lin:2017,Lee:2018}. 



Another family of 3D pose estimators relies on 2D pose estimators by first predicting 2D key joint positions in image space~\cite{Newell:2016,Wei:2016,Papandreou:2017,Cao:2018,Chen:2018}, which are subsequently lifted to 3D either by regression~\cite{Martinez:2017,Moreno-Noguer:2017} or model fitting~\cite{Chen:2017}. The latter methods, which are the most popular, are benefited by the availability of large-scale datasets and perform a k-nearest neighbor search for a predicted set of 2D keypoints over a dictionary of 2D skeletons for which the 3D skeleton is available, and then simply output the corresponding 3D pose~\cite{Yiannakides:2019,Zhou:2017,Tekin:2017}. Some extensions integrate joint constraints to overcome the ambiguity in transferring 2D human poses into their 3D pose representations~\cite{Akhter:2015,Mehta:2019:Xnect}, enforce priors about bone lengths, and projection consistency with the 2D ground truth~\cite{Ramakrishna:2012,Brau:2016,Dabral:2018,Wang:2019}, or use multi-view cameras to improve supervision~\cite{Rhodin:2018_CVPR}. Others use adversarial training to generate anthropometrically feasible poses~\cite{Yang:2018}. In general, these methods outperform those based on direct estimation, since they benefit from intermediate supervision, which can be trained on images in the wild. Indeed, the state-of-the-art results for 3D pose estimation are achieved by a recent two-stage method of Pavllo et al.~\shortcite{Pavllo:2019}. 


However, a major limitation of these pose estimation methods, is that the positional representation can only describe a small set of the full human motion articulation, which highly complex and dynamic. 
Moreover, since motion is not assigned to a character model to enforce kinematic constraints, resulting temporal bone length may be violated. \dl{Some works use reference objects in the scene with known position and scale, or other distinct parts of the human body (e.g., heads), to better estimate the person's height~\cite{Gunel:2018,Bieler:2019}. More recently, body regression models have been employed to enforce some form of skeleton constraints~\cite{Mehta:2019:Xnect,Yoshiyasu:2018}.}
Our goal is to enable a complete representation of motion, which consists of a skeleton and joint rotations.

In addition, positional representation allows ambiguities regarding the geometry and orientation of the limb segments as illustrated in Figure~\ref{fig:position_ambiguity}). A full motion representation requires reconstruction of joint rotations. Converting a positional representation into 3D joint rotations is rather challenging, since this problem is highly under-constrained~\cite{Dong:2017}. 
 Mehta~\etal~\shortcite{Mehta:2017} fit a rigged skeleton model over time to estimated 2D and 3D joint locations, and use IK as a post-process to recover 3D rotations of each joint. In contrast, our network directly extracts 3D joint rotations from 2D joint positions, producing the same format as that of commercial motion capture systems. The sequence of 3D joint rotations refers to a single 3D skeleton, thus ensuring that bone lengths remain constant over time, and learning joint quaternions directly from the data results in a natural motion without assuming an underlying model or employing IK solvers.

\paragraph{\textbf{3D shape recovery:}}
A parallel, related line of research aims at recovering full 3D shapes of people from images~\cite{huang2017towards,Kolotouros:2019:CVPR}, using mainly \dl{statistical body shape (SMPL)} models~\cite{Loper:2015}. These approaches are typically based on parameterized 3D meshes and regression, and focus less on pose accuracy~\cite{Anguelov:2005,Loper:2015}. More recently, a number of deep convolutional methods have been proposed that map all human pixels of an RGB image to the 3D surface of a human body~\cite{Lassner:2017,Guler:2018,Alldieck:2018,Zheng:2019}. Next, in order to extract the 3D human pose from silhouettes, they fit a body shape model to the 2D joints, and minimize an objective function that penalizes the error between the projected 3D model joints and detected 2D joints~\cite{Kanazawa:2018,Pavlakos:2018,Zhou:2018, Kolotouros:2019:ICCV}. \dl{As a consequence, they predict the joint rotations as a function of the surface}. 
\dl{Moreover, additional steps and optimizations are required to ensure temporal consistency on the geometry mesh~\cite{Bogo:2016,Guler:2019,Xu:2019:DenseRac,Xu:2018,Arnab:2019}, including, in some cases, multi-view cameras to learn a geometry-aware representation of human shape, where the 3D pose is estimated through direct regression~\cite{Rhodin_2018_ECCV}, or the employment of networks that can learn and predict the human dynamics~\cite{Kanazawa:2019:Dynamics,Zhang2019:Dynamics}.} 
\dl{However, these works do not guarantee a consistent skeleton across different frames. Even though temporal coherence of the geometric model is encouraged, phenomena such as bone stretching may appear. In order to extract a single kinematic skeleton from such a geometric model, rigging should be applied. In contrast, our work aims to reconstruct a set of joint rotations along with a single kinematic skeleton in an end-to-end, data-driven fashion.}


\dl{Unlike these methods, we utilize a single network that supports direct regression of 3D joint rotations, and at the same time ensures the temporal consistency of the skeleton, solving both problems end-to-end. More specifically,} instead of fitting a skeleton to detect the 3D human pose, we integrate kinematics into the neural network, similarly to~\cite{Zhou:2016,Villegas:2018,Kanazawa:2018,Pavllo:2018:Quaternet}, \dl{where joint angles are recovered by learning the rotational data directly}. 
\dl{However, Zhou et al.~employ loss only on joint locations, and the joint angles are estimated as a byproduct, while Kanazawa et al.~recover the joint rotation angles as part of the parameters for the SMPL body model using unpaired adversarial training, with a separate discriminator for each joint. On the other hand, Pavllo et al.~\shortcite{Pavllo:2018:Quaternet} focus on human motion prediction by regressing the joint rotations, applying forward kinematics, and then employing a loss on joint positions.} In contrast, our network learns to output temporally coherent joint angles with natural velocities using training data from available motion capture datasets, and is trained on motion sequences, as opposed to individual frames. In addition, the structure of our network enforces a single consistent skeleton, which can be directly converted into an animation file without the need to apply IK or temporal smoothing~\cite{peng2018sfv,Arnab:2019}.


\section{Motion Reconstruction}
At the crux of our approach lies a deep neural network, MotioNet, whose unique structure is inspired by the common working methodology in the animation field. MotioNet enables extraction of 3D poses with natural, data-driven, rotational information, which can be directly converted into motion file formats used in animation platforms and game engines. 

In practice, our network learns to map 2D joint positions, extracted from an input video, into two separate components: (i) a single, symmetric, skeleton, represented by its bone lengths; (ii) a dynamic, skeleton-independent, sequence of: joint rotations, global root positions and foot contact labels. These two parts constitute a complete, global, description of motion, requiring no further processing, or IK, to obtain a full 3D animation. 

To train our network, we leverage motion capture datasets that comprise temporal sequences of various motions performed by different subjects, where each motion is represented by a single skeleton and a temporal sequence of 3D joint rotations and global positions.
During training, the 3D motion is projected onto 2D planes from arbitrary view angles, and the network is trained to recover the aforementioned components from the projected sequence. The foot contact labels are extracted from the captured motions by examining the distance of the foot joints from the ground and their velocity. \dl{During test time, the predicted labels can used to constraint the position of the joints, via IK optimization}

\dl{Our main technical contribution is a new 2-branch network that reconstructs the dynamic properties of the motion -- joint rotations, global root positions, foot contact labels, separately from the static ones -- a single kinematic skeleton. Our network consists of various components, some of which were previously proposed by other methods, while others are new. The forward kinematics layer that we use was originally proposed by Villegas et al. \shortcite{Villegas:2018} in the context of motion retargeting. However, we are the first to use it for pose estimation. In addition, Kanazawa et al.~\shortcite{Kanazawa:2018} already applied a discriminator to joint rotation angles, but did so in a per-frame fashion and using absolute angle values. In contrast, our discriminator judges the realism of temporal sequences of angular velocities. Lastly, Pavllo et al.~\shortcite{Pavllo:2019} have also used 1D temporal convolution in the context of 3D pose estimation. However, their network is trained to lift 2D joint positions into 3D, while ours is trained to convert joint positions to rotations.}


\begin{figure*}
	\includegraphics[width=\linewidth]{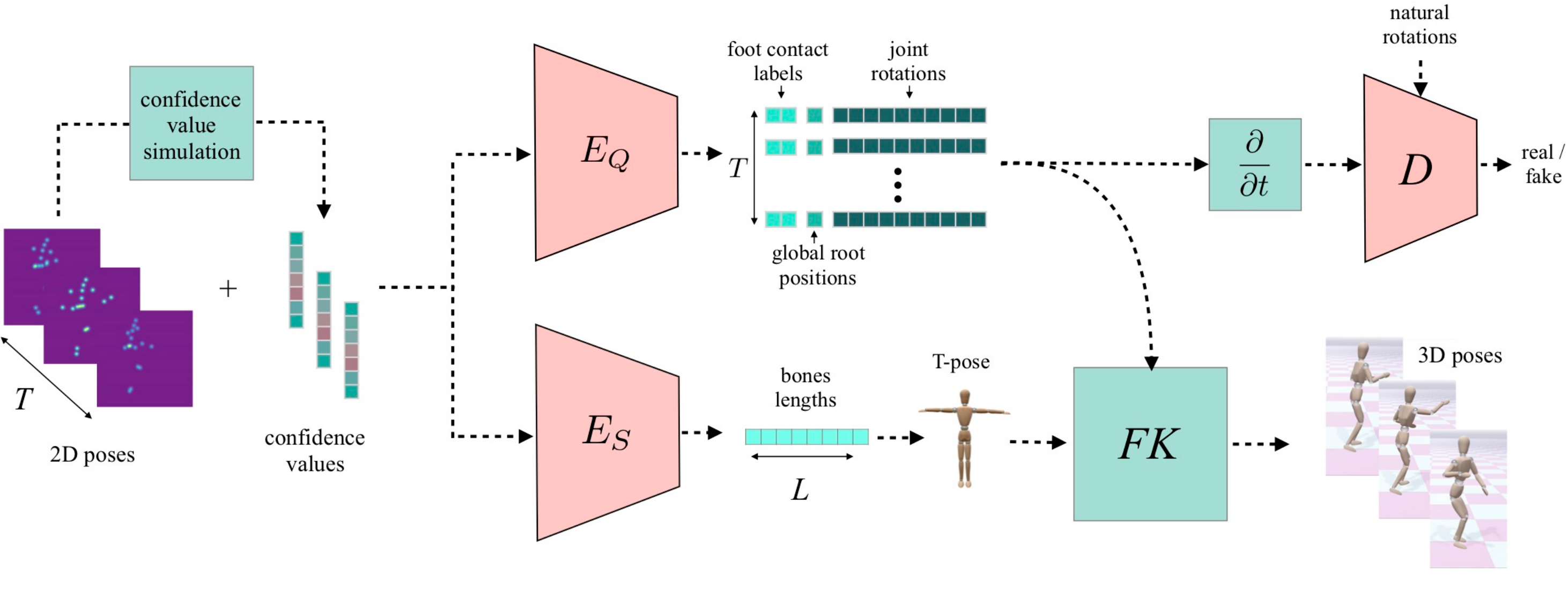}
	\caption{Our framework receives 2D joint positions along with per-joint confidence values, which are simulated based on empirical experiments from real videos. It extracts per-frame joint rotations and global root positions along with foot contact labels, and a static (duration-independent) skeleton, using two encoders, $E_Q$ and $E_S$. The extracted rotations are fed into a discriminator $D$ that is trained to tune the temporal differences of the rotation angles to mimic the distribution of natural rotations, using adversarial training. In addition, the rotations and the static feature that is converted to a ``T-pose'', are fed into the forward kinematic layer, $FK$, that extracts 3D joint positions, which are compared to a ground truth.}
	\label{fig:high_level_scheme}
\end{figure*}

\subsection{Network Architecture}
\label{subsec:network}
Let $\bbp_{s,q,r}\in\mathbb{R}^{T\times 3J}$ denote a temporal sequence of 3D joint positions generated by a skeleton $\bs\in\mathbb{R}^L$ with joint rotations $\bq\in\mathbb{R}^{T\times QJ}$ and global root positions $\br\in\mathbb{R}^{T\times 3}$, where $L$ is the number of limbs, $T$ is the temporal length of the sequence, $J$ is the number of joints, and $Q$ is the size of the rotations representation vector, which will be discussed later. 


Our goal is to train a deep neural network to decompose sequences of projections of 3D joints, $C(\bbp_{s,q,r}; c)\in\mathbb{R}^{T\times 2J}$ ($C$ is a perspective projection operator, which projects the 3D joint positions into 2D using camera parameters $c\in\mathcal{C}$), into four attributes: $\tilde{\bq}\in\mathbb{R}^{T\times QJ}$, that captures the dynamic, rotational information of the motion, $\tilde{\bs}\in\mathbb{R}^L$, that describes the single, consistent, skeleton, $\tilde{\br}\in\mathbb{R}^{T\times 3}$ that estimates the global positions of the root along time, and $\tilde{\bff}\in\{0,1\}^{T\times 2}$ that predicts whether each of the two feet touches the ground in each frame. The first three attributes can be then combined via forward kinematics to estimate the global 3D pose sequence, $\tilde{\bbp}_{\tilde{s},\tilde{q},\tilde{r}}\in\mathbb{R}^{T\times 3J}$, specified by joint positions.

A high level diagram of our approach is shown in Figure~\ref{fig:high_level_scheme}. During training, each data sample, $C(\bbp_{s,q,r}; c)\in\mathbb{R}^{T\times 2J}$, is fed, in parallel, into two networks, $E_S$ and $E_Q$. The task of $E_s$ is to estimate the single skeleton
\begin{equation}
\tilde{\bs} = E_S(C(\bbp_{s,q,r}; c)),
\end{equation}
while $E_Q$ aims to estimate rotations, global root positions, and foot contact labels via
\begin{equation}
  \tilde{\bq}, \tilde{\br} ,\tilde{\bff} = E_Q(C(\bbp_{s,q,r}; c))],
\end{equation}
respectively. 
We train a single network to estimate all the three dynamic attributes due to the strong correlation between them (e.g., the global root position of a running human, is temporally correlated with the rotations of the limbs, and the timing of contact with the ground).

Since the rotations, global positions and contact labels are dynamic (change per frame), we design $E_Q$ to retain temporal information. In practice, it employs one-dimensional convolution layers over the time axis, yielding a temporal set of joint rotations. In contrast, the skeleton is a static attribute, and $E_S$ employs adaptive pooling to collapse the temporal axis, resulting in a fixed size vector of bone lengths, regardless the input dimensions, as illustrated in Figure~\ref{fig:layers}. 

\begin{figure}
	\centering
	\includegraphics[width=1.0\columnwidth]{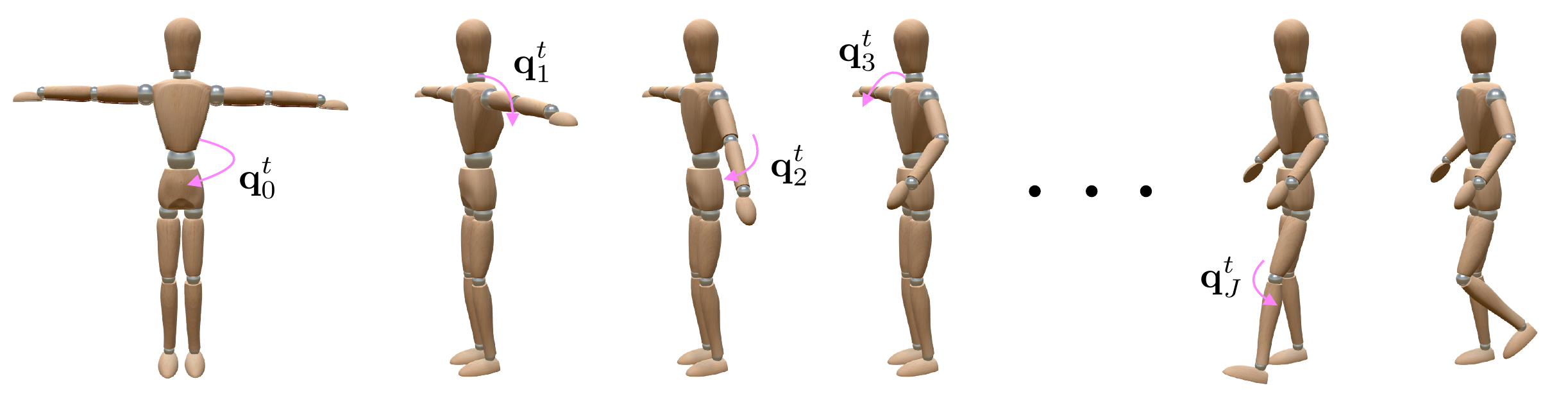}
	\caption{Our network applies forward kinematics on a ``T-pose'' skeleton by successively rotating the limbs from the root to the end-effectors.}
	\label{fig:FK}
\end{figure}

\paragraph{Forward Kinematics}
In order to convert $\tilde{\bs}$, $\tilde{\bq}$ and $\tilde{\br}$ into 3D joint positions, we use Forward Kinematics (FK), which refers to the process of computing the joint positions, given the joint rotations and the initial skeleton. FK is performed by successively rotating the limbs of an input skeleton tree starting from the root joint and ending in the leaf joints (see Figure \ref{fig:FK}).
Since these are linear operations, we can integrate them into the network as a differential layer \cite{Villegas:2018} $FK$, which allows back propagation. In practice, given the estimated bone lengths, $\tilde{\bs}$, we first reconstruct a ``T-pose'', $\tilde{\bs}_{\textrm{init}}\in\mathbb{R}^
{3J}$, 
by positioning the joints in 3D coordinates. This is done by placing the root at the origin, while the other joints are placed on the $XY$ plane, based on the offsets specified in $\tilde{\bs}$ (as demonstrated by the left most character in Figure~\ref{fig:FK}). Then, at each time step $t$, the $FK$ layer is applied to $\tilde{\bs}_{\textrm{init}}$, using the joint rotations of the $t$-th frame, $\tilde{\bq}^{t}\in\mathbb{R}^{4J}$ to obtain
\begin{equation}
\tilde{\bbp}^{t}_{\tilde{s}, \tilde{q}, \tilde{r}=0} = FK(\tilde{\bs}_{\textrm{init}}, \tilde{\bq}^{t}),
\end{equation}
where $\tilde{\bbp}^{t}_{\tilde{s}, \tilde{q}, \tilde{r}=0}$ is the local 3D pose (root positioned in the origin) at frame $t$.
Rotations are applied successively to each joint starting from the root and ending at the leaf joints,  $\tilde{\bbp}^{t}_n = \tilde{\bbp}^{t}_{\textrm{parent}(n)} + R^{t}_{n}\tilde{\bs}_{n}$, where $\tilde{\bbp}^{t}_n\in\mathbb{R}^3$ is the position of the $n$-th joint at time $t$, $\tilde{\bbp}^{t}_{\textrm{parent}(n)}\in\mathbb{R}^3$ is the position of its parent, $R_n\in\mathbb{R}^{3\times 3}$ is the rotation matrix of the $n$-th joint, and $\tilde{\bs}_{n}\in\mathbb{R}^3$ is the 3D offset of the $n$-th joint relative to its parent in $\tilde{\bs}_{\textrm{init}}$.
Finally, given the local pose of the 3D skeleton, we add the global translation to obtain the full estimated 3D pose in frame $t$,
\begin{equation}
\tilde{\bbp}_{\tilde{s},\tilde{q},\tilde{r}}^{t} = FK(\tilde{\bs}_{\textrm{init}}, \tilde{\bq}^t) + \tilde{\br}^{t}.
\end{equation}
Since FK consists of differential operators, it can be integrated into the network as a layer through which back-propagation may be applied.
The FK process for a given time step is illustrated in Figure~\ref{fig:FK}.


%
%
%
%

\subsection{Training and Loss}
\label{subsec:loss}
To train our network, we leverage publicly available motion capture datasets described in more detail in Section \ref{subsec:dataset}. We use a combination of full self-supervised and partial self-supervised training. For the fully supervised training we use captured human motions for which the full, rotational and positional 3D joint data is available, while the partially supervised training only makes use of 3D joint positions (positional representation). In both cases, the 2D inputs are generated automatically by projecting the 3D training data, thus the paired training is essentially self-supervised.




The loss function that we use consists of four components: skeleton loss $\Loss_{\text{S}}$, global root position loss $\Loss_{\text{R}}$, joint position loss $\Loss_{\text{P}}$, and rotations GAN loss $\Loss_{\text{Q_{GAN}}}$. These components are described in more detail below.

\paragraph{Skeleton Loss}
Let $\bbp_{s,q,r} \in \mathcal{P}$ denote a 3D motion sequence in our dataset $\mathcal{P}$, and $C(\bbp_{s,q,r}, c_i)$ denote its 2D projection using a camera $c_i\in\mathcal{C}$. The skeleton loss $\Loss_{\text{S}}$ ensures that the encoder $E_S$ correctly extracts the skeleton $\bs$:
\begin{equation}
\Loss_{\text{S}} = \bbe_{\bbp_{q,s,r}\sim \pp, c_i\sim \cc}\left[ \Vert E_S(C(\bbp_{s,q,r}, c_i))-\bs\Vert^2 \right].
\end{equation}
When perspective projection is used, the 3D skeleton can be reconstructed only up to a scale factor. In order to avoid this scale ambiguity, we globally rescale all the 3D skeletons in the data such that their average bone length is equal to $1$. At inference time, if the real spine length of the performer is known, the reconstructed 3D pose can be rescaled into a real metric space.

\paragraph{Joint Position Loss}
This is our main loss, designed to ensure that the joints in the extracted poses are placed in the right 3D positions.
Unlike naive supervised reconstruction of 3D joint positions, in this case, the positions are constrained by the geometric structure of the reconstructed skeleton, which leads to smaller, correlated, search spaces.
Although the first part of the network ($E_S$ and $E_Q$) is supposed to output the full, positional and rotational description of the motion, the joint position loss is essential since a direct loss over the joint rotations is not applied. In addition, since errors can be accumulated during FK from the root to the end-effectors, the FK-reconstructed joint positions, per frame, tune the joint rotations such that the reconstructed 3D joint position match the original ground truth ones:
\begin{equation}
\Loss_{\text{P}} = \bbe_{\bbp_{q,s,r}\sim \pp, c_i\sim \cc}\left[ \Vert FK(\tilde{\bs}_{\textrm{init}}, \tilde{\bq})-\bbp_{s,q,r=0}\Vert^2 \right].
\end{equation}
Since human observers are particularly sensitive to the motion of end effectors in animations, we split the joints into two groups and penalize the end-effectors part, $\Loss_{\text{P}_{EE}}$ with a different weight ($\lambda_{P_{EE}}$).

\paragraph{Adversarial Rotation Loss}

Applying a loss (only) over the skeleton and joint positions, trains the network to reconstruct correct positions, without imposing constraints on the rotations. 
Due to the inherent ambiguity that exists in IK, the resulting rotations and their velocities might be unnatural, which can lead to distortions when skinning is applied to the skeleton.
Although a part of the data contains 3D rotations that were captured in motion capture systems, a direct loss applied to the rotations may not converge. Due to the fact that the T-poses for different samples in the dataset are not aligned, namely, two similar poses might be represented by different rotations, thus, a direct loss on the rotations can not be applied, unless, the entire set is retargeted to share the T-pose, which is a time consuming operation.
Due to the potential difference between the absolute values of rotations that represent the same pose, our network is trained to output rotations with natural velocities distribution using adversarial training. The idea is to focus on the temporal differences of joint rotations rather than their absolute values. For example, if a specific joint is constrained to move only in a specific direction we want to follow that rule, in a data-driven manner. In order to incorporate the adversarial loss we add $J$ discriminators, $D_j$, where each is aimed to distinguish between temporal differences of the output rotations and real rotations from the dataset for a specific joint,
\begin{eqnarray}
\Loss_{\text{Q_{GAN}}_j} & = &\bbe_{q \sim \qq}
\left[ \Vert D_j(\Delta_t q_j) \Vert^2 \right] \\ \nonumber
&+& \bbe_{\bbp_{q,s,r}\sim \pp, c_i\sim \cc} \left[ \Vert 1-D_j(\Delta_t E_Q(C(\bbp_{s,q,r}, c_i))_{q_j}\Vert^2\right],
\end{eqnarray}
where $E_Q(\cdot)_{q_j}$ denotes the rotations of the $j$th joint of the output of $E_Q$, $\qq$ represents the distribution of natural joint angles that are taken from the dataset, and $\Delta_t$ stands for temporal differences. The full architecture of each components can be found in the Appendix~\ref{appendix}.

Note that due to the averaging behavior of the temporal convolutions and the temporal continuity of the input sequence, the estimated joint rotations are encouraged to be smooth along the temporal axis, resulting in smooth animations.

\paragraph{Reference T-pose}
Since our rotations are guided by velocities, their absolute values might be unnatural and erroneous. These errors can be observed when skinning is applied to the underlying skeleton. In order to let the network learn a correct reference point, in each epoch, we train the network with a few T-poses (T-pose per character), and require that their reconstructed rotations will be zero using a direct loss over the output rotations.

\paragraph{Global Root Position Loss}
In order to avoid the learning of unnecessary information in the reconstruction of the global root position in frame $t$, $\br^t=(X_r^t,Y_r^t,Z_r^t)$, we estimate only the depth parameter $Z_r^t$ and then derive the global root position by
\begin{equation}
\left( X_r^t,Y_r^t,Z_r^t\right)  = \left( \frac{Z_r^t}{f}x_r^t,\frac{Z_r^t}{f}y_r^t,Z_r^t\right),
\end{equation}
where $(x_r,y_r)$ is the root position in the projected 2D input sequence, and $f=1$ is the assumed focal length of the camera model. Then the loss is given by
\begin{equation}
\Loss_{\text{R}} = \bbe_{\bbp_{q,s,r}\sim \pp, c_i\sim \cc}\left[ \Vert E_Q(C(\bbp_{s,q,r}, c_i))_r - Z_r\Vert^2 \right],
\end{equation}
where $E_Q(\cdot)_r$ denotes the root position part of the output of $E_Q$.

\paragraph{Foot Contact Loss}
In order to handle foot sliding artifacts we supervisedly train the network to output $\tilde{\bff}$ that predicts whether each foot is in contact with the ground in each frame via
\begin{equation}
\Loss_{\text{F}} = \bbe_{\bbp_{q,s,r}\sim \pp, c_i\sim \cc}\left[ \Vert E_Q(C(\bbp_{s,q,r}, c_i))_f - \bff\Vert^2 \right],
\end{equation}
where $E_Q(\cdot)_f$ denotes the foot contact label part, $\tilde{\bff}\in\{0,1\}^{T\times 2}$, of the output of $E_Q$.
Based on that binary GT vector, $\bff$, another loss, that encourages the velocity of foot positions to be zero during contact frames, is added
\begin{equation}
\Loss_{\text{FC}} = \bbe_{\bbp_{q,s,r}\sim \pp, c_i\sim \cc}\left[ \Vert \bff_i \sum_j \Delta_t FK(\tilde{\bs}_{\textrm{init}}, \tilde{\bq})_{f_i} \Vert^2 \right],
\end{equation}
where $FK(\cdot,\cdot)_{f_i}\in\mathbb{R}^{T\times 3}$ and $\bff_i$ denote the positions and the contact labels of one of the feet joints ($i\in \mathit{left}, \mathit{right}$), and $\sum_j$ sums the components along the coordinate axis.
The contribution of this component is evaluated in the ablation study in Section~\ref{subsec:ablation}.


Summing the four terms, we obtain our total loss:
\begin{equation}
\Loss_{\text{tot}} = \Loss_{\text{P}} +
\lambda_{\text{S}} \Loss_{\text{S}} +
\lambda_{\text{Q}}\sum_{j\in J}\Loss_{\text{Q_{GAN}}_j} +
\lambda_{\text{R}}\Loss_{\text{R}} +
\lambda_{\text{F}}\Loss_{\text{P}_{F}} +
\lambda_{\text{FC}}\Loss_{\text{P}_{FC}},
\end{equation}
where in all of our experiments $\lambda_{\text{S}}=0.1$, $\lambda_{\text{Q}}=1$, $\lambda_{\text{R}}=1.3$,  $\lambda_{\text{F}}=0.5$, $\lambda_{\text{FC}}=0.5$.

\subsection{Motion Dataset}
\label{subsec:dataset}

Our training data is constructed from the CMU motion capture dataset \cite{CMU:mocap}, and from the Human3.6M human pose dataset \cite{Ionescu:2014}.
The CMU dataset consists of 2605 captured motions performed by 144 subjects. The motions include elementary actions (walking, jumping, kicking, etc.), and various dancing moves (samba, salsa, etc.).
The Human3.6M dataset consists of over 3.6 million 3D human poses with corresponding images, viewed from four different angles using synchronized cameras. The motions are performed by 11 actors and include 17 everyday scenarios, such as eating, smoking, talking on the phone, walking, etc.
The actors in both datasets feature a variety of body shapes and skeleton proportions. 

\dl{In order to train the network to extract foot contact labels in a supervised fashion, we must first extract ``ground truth'' labels from the dataset. Specifically, we consider a foot to be in contact if its height above the ground is below 20mm, and its average velocity magnitude in the surrounding 5-frame window is below 1mm/frame. The height of the ground is approximated by the average of the lowest 20-percentile of all the foot heights in the sequence.}


The actual input data samples fed into the network are generated by projecting the sequences of 3D joint positions (represented in camera coordinates space) onto different camera view angles, using perspective projection.
As a result, we obtain a rich dataset with ground truth, which demonstrates how skeletons of different characters, performing various motions, appear from different views when projected into 2D.

In test time, when videos in the wild are used, we use the method of Cao \etal~\shortcite{Cao:2018} to extract 16 joints (root, neck, thorax, head, shoulders, hips, knees, feet, elbows, and hands). These joints exist in the aforementioned datasets as well. In addition, a spine joint which exists only in the 3D datasets is artificially added (calculated as the 2D spatial average between the root and the neck joint).

\paragraph{Preprocessing}
In order to normalize the 2D input data we apply two operations: (i) Subtract the root position from every joint location (in all the frames), (ii) Subtracting the mean joint position and divide it by the standard deviation, while the mean and standard deviation are empirically calculated per joint over the entire dataset. Since global information is discarded from the normalized local representation, we append to it the global 2D velocity (per-frame).

\subsection{Implementation Details}
\begin{figure}
\centering
	\includegraphics[width=\linewidth]{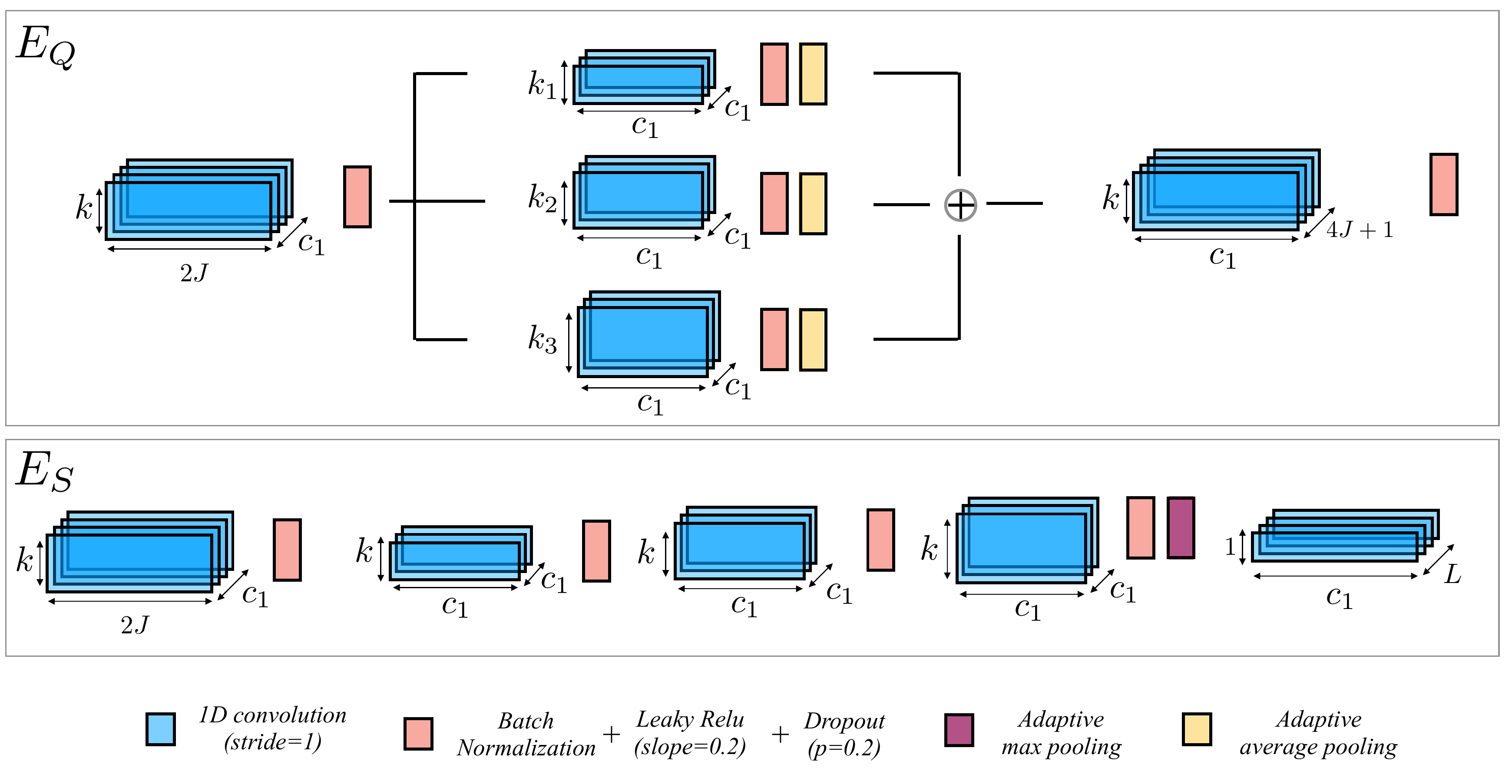}
	\caption{
	Our network contains two encoders,  $E_Q$ which generates a temporal set of joint rotations, global positions and foot contact labels using parallel convolutions, and $E_S$ which outputs a static attribute that represents the skeleton, using an adaptive pooling layer that collapses the temporal axis.}
	\label{fig:layers}
\end{figure}

In practice, our implementation consists of two encoders, $E_Q$ and $E_S$, a forward kinematics layer $FK$, and a discriminator $D$. The layers and the dimensions of the two encoders are shown in Figure~\ref{fig:layers}, while the $FK$ layer is described by Villegas et al.~\shortcite{Villegas:2018}.
For a detailed description of the parameters of each layer, please refer to the appendix.

In the $E_Q$ encoder there are five 1D convolution layers. The first layer expands the number of channels, and the result is then fed in parallel into three convolution layers with kernels of three different sizes. This is inspired by the inception module from GoogLeNet.
The three results are then summed together after adaptive pooling and fed into a final layer that generates the temporal sequence of quaternions and root depths.

The $E_S$ encoder uses a simple sequence of 1D convolution layers. Again, the first layer expands the number of channels, and the last layer outputs a vector of bone length, which implicitly enforces symmetry, since each pair of symmetric limbs is encoded by the same coordinate.

The discriminator $D$ also uses a sequence of 1D convolution layers. There is a per-joint convolution layer which is summarized into a fully connected layer that dictates whether the rotational velocity is real or fake.


\section{Supporting Videos in the Wild}
\label{sec:wild}
During training, the input to our network is produced by projecting mostly clean motion captured data from 3D to a 2D image plane. However, at test time, the network should be able to cope with 2D joint positions extracted from videos in the wild, using the OpenPose library \cite{Cao:2018}. In order to overcome the gap between the distribution of the training inputs and the test-time ones, we employ confidence values and augmentation.

\paragraph{Confidence Values}
During test time, when 2D joint positions are extracted from videos in-the-wild, there might positional errors or even missing joints due to occlusion.
In order to simulate such input characteristics during the training, we leverage the output structure that state-of-the-art 2D pose estimators provide and attach for each joint a value $c_n \in [0,1]$, which specifies the confidence in the estimation of that joint's 2D position. For a completely missing joint the confidence $c_n$ is set to 0.

\begin{figure}
\centering
	\includegraphics[width=\linewidth]{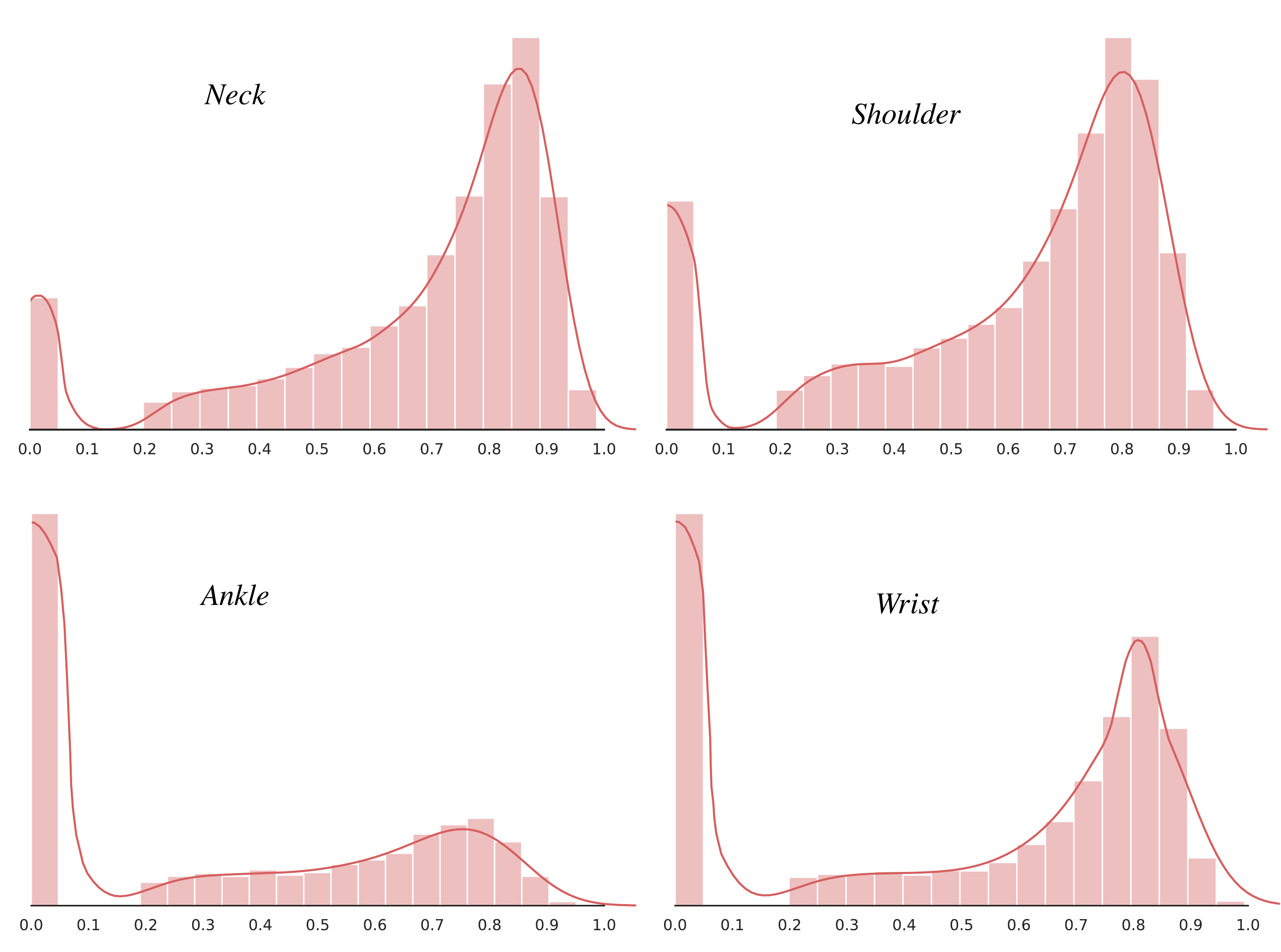}
	\caption{Modeling the distribution of joint confidence values (continuous red line) using empirical distribution of confidence values (bins), extracted from videos in the wild.
	}
	\label{fig:conf_dist}
\end{figure}

Thus, in practice, we concatenate another vector $\bc\in\mathbb{R}^{J\times T}$ to our input, both during training and at test time. At test time the values of $\bc$ are extracted from the output of Cao \etal~\shortcite{Cao:2018}, while during training we randomly draw the values, based on an empirically determined distribution. The distribution was determined using the confidence values that were extracted from a diverse set of videos in-the-wild taken from~\cite{Johnson10}. We calculate the histogram of each joint separately, and model each distribution as a sum of two parts: a delta distribution around zero (with probability of $\delta_j$) and a Gaussian distribution $(\mu_j, \sigma_j)$ clipped at 1 (probability of $1-\delta_j$). Figure~\ref{fig:conf_dist} plots the normalized histograms of the confidence values for 4 different joints and the per-joint parameters ($\delta_j, \mu_j, \sigma_j$) are listed in the appendix. It may be seen that joints that are closer to the end-effectors tend to be occluded more frequently.

Once a confidence value $c_n>0$ is drawn, we compute a spatial displacement, which is applied as a noise to the 2D coordinate. The direction of the displacement is drawn uniformly from $[0^{\circ}, 360^{\circ}]$, while the magnitude is drawn from $[0, \beta(1-c_n)]$ for $\beta=6$. For $c_n=0$ we set the value of the coordinate to $(0,0)$. Thus, similarly to denoising autoencoders, our network is trained to be robust to inaccurately estimated, or even missing joint locations.

\paragraph{Augmentation}
To enrich the observed samples, we apply data augmentation in two different ways: 
\begin{enumerate}
\item Clip length: since our model is fully convolutional, input motion clips with varying lengths can be forwarded through the same network. However, due to the use of batches a fixed temporal length can improve training time. Thus, we randomly pick the temporal length ($T$) of the samples in the batch, per iteration, to be an integer in the range of 60 to 200. This augmentation step enhances the disentanglement of the static parameters and the temporal length of the sequences.

\item Camera augmentation: since we work in the 3D camera coordinate space, our projections are augmented by modifying the depth of the 3D character (global translation along the $z$ axis) and its orientation (global rotation of the root).
This operation is important since it trains the network to map similar poses in different 2D scales into the same (local) 3D parameters (rotations and bones length), thus extending the distribution of the inputs of the training data.

\item Flipping: we left-right flip the 3D joint positions (camera coordinate space) of the samples, before the projections, to obtain augmented poses with $\tilde{P}_j^r=\left( -(P^l_j)_x, (P^l_j)_y, (P^l_j)_z \right)$, $\tilde{P}^l=\left( -(P^r_j)_x, (P^r_j)_y, (P^r_j)_z\right) $, were $P^r_j$ and $P^l_j$ are the left and right positions of a symmetric joint $j$ (e.g, left and right shoulder). For single joints without symmetry (root, neck, etc.) we use  $\tilde{P}_j=\left( -(P_j)_x, (P_j)_y, (P_j)_z \right)$.

\end{enumerate}


\section{Results and Evaluation}
\label{sec:results}

\begin{figure}
\includegraphics[width=0.98\columnwidth]{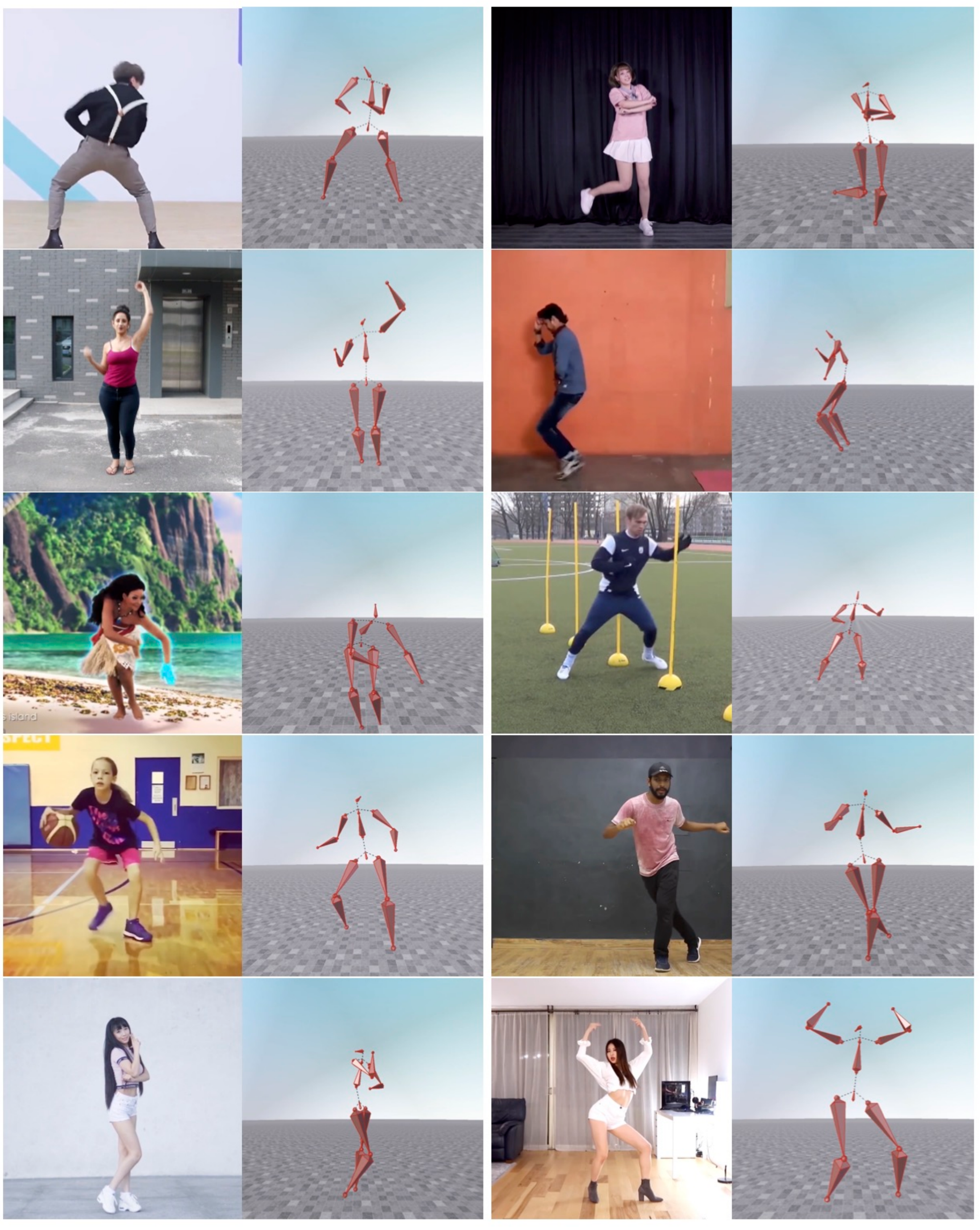}
\caption{Examples of motions reconstructed by MotioNet from videos in-the-wild. MotioNet can reconstruct a variety of human motions, while being robust to partial occlusions and tolerant to different body types. Frames from the video are on the left, and the reconstructed pose is on the right.}
\label{fig:gallery}
\end{figure}

In this section, we show 3D human motion reconstruction results obtained by MotioNet, conduct experiments and ablation studies to analyze the performance of various components in our framework and present comparisons to state-of-the-art 3D human motion estimation techniques. Figure \ref{fig:gallery} shows a gallery of results which demonstrate the ability of MotioNet to extract a variety of human motions from videos in the wild, while being robust to partial occlusions and tolerant to different body types. In order to produce the animations, we convert the tensors produced by our network into BVH files, and use the Blender software to read these files and render the videos, without any further post processing. The quality of the resulting animations may be examined in the supplementary video.

\subsection{Comparison to Other Methods}
\label{subsec:comp_pos}
We first compare the accuracy of the reconstruction of  joint positions by our approach to a variety of existing methods. 

\paragraph{Quantitative comparison} A quantitative comparison is reported in Table~\ref{tab:quant_human36}, which reports the mean per joint position error (MPJPE):
\begin{equation}
E(\tilde{\bbp},\bbp) =\frac{1}{J}\sum_{j=1}^J \Vert\tilde{\bbp}_j - \bbp_j\Vert.
\label{eq:mpjpe}
\end{equation}


\dl{We evaluated our method on two datasets. The first one is Human3.6M dataset \cite{Ionescu:2014} which adopts a 17-joint skeleton representation. For this dataset, the network was trained on five subjects (S1, S5, S6, S7, S8), and tested on two (S9 and S11), using a single model that was trained for all actions. The second dataset is HumanEva-I \cite{Sigal:2009} which adopts a 15-joint skeleton representation, including three different subjects. We evaluated our system on two actions (Walk, Jog) by training a single model. Note that similarly to other approaches, the evaluation with this dataset is computed after performing a rigid alignment with the ground truth including scale, rotation, and translation (a.k.a P-MPJPE).}

\dl{In order to apply a fair comparison we did not simulate the confidence map, and used 2D detected key-points during both training and test time. We followed Pavllo et al.~\shortcite{Pavllo:2019} and used the cascaded pyramid network (CPN)~\cite{chen2018cascaded} to extract 2d poses. The bounding box required by CPN is extracted by Mask-RCNN~\cite{he2017mask} and the 2d detector is fine-tuned by 2d ground truth samples from Human3.6M.}

As may be seen from the results in Table~\ref{tab:quant_human36} and Table~\ref{tab:quant_humaneva}, while our method does not achieve the highest accuracy for joint positions, it is usually among the top 2--3 methods. This is despite the fact that our method aims at a slightly different task. Furthermore, note that enforcing a single underlying skeleton may introduce some errors into the reconstructed joint positions due to inaccuracies (even small ones) in the estimated bone lengths. In order to measure the effectiveness of the the rotations branch and eliminate the effect of errors in the skeleton reconstruction, we conduct an experiment that uses the ground truth (GT) skeleton instead of the reconstructed one. In this experiment only $E_Q$ was trained, while $FK$ received the GT skeleton. The result reported in Table~\ref{tab:quant_human36} shows that if the exact skeleton is known, the accuracy of poses can be even higher.

\dl{In order to demonstrate the effectiveness of our dynamic encoder $E_Q$, we replaced its structure by the dilated 1-D convolution architecture proposed by Pavllo et al.~\shortcite{Pavllo:2019}, originally designed to output positions. The results which demonstrate that our architecture fits better to the prediction of rotations are also depicted in Table~\ref{tab:quant_human36}.}

\begin{table*}
\resizebox{\textwidth}{!}{

\begin{tabular}{c|ccccccccccccccc|c}
&Dir. & Disc.&  Eat & Greet & Phone & Photo & Pose & Purch. &Sit& SitD.& Smoke &Wait& WalkD.& Walk& WalkT. & avg.(mm)\\
Mehta \etal~\shortcite{Mehta:2017} & 62.6& 78.1& 63.4& 72.5& 88.3& 93.8&  63.1& 74.8& 106.6& 138.7& 78.8& 73.9& 82.0& 55.8&  59.6& 80.5\\
Pavlakos \etal~\shortcite{Pavlakos:2017} &  67.4& 71.9& 66.7& 69.1& 72.0& 77.0& 65.0& 68.3& 83.7& 96.5& 71.7& 65.8& 74.9& 59.1& 63.2& 71.9\\
Martinez \etal~\shortcite{Martinez:2017} & 51.8& 56.2& 58.1& 59.0& 69.5& 78.4& 55.2& 58.1& 74.0& 94.6& 62.3& 59.1& 65.1& 49.5& 52.4& 62.9\\
\dl{Yoshiyasu \etal~\shortcite{Yoshiyasu:2018}} & 63.3& 71.6& 61.4& 70.4 & 69.9& 83.2& 63.0& 68.8& 76.8& 98.9& 68.2& 67.5& 73.7& 57.7& 57.1& 69.9\\
Fang \etal~\shortcite{Fang:2018} &50.1& 54.3& 57.0& 57.1& 66.6& 73.3& 53.4& 55.7& 72.8& 88.6& 60.3& 57.7& 62.7& 47.5& 50.6& 60.4\\
Yang \etal~\shortcite{Yang:2018}($+$) &51.5& 58.9& 50.4& 57.0& 62.1& 65.4& 49.8& 52.7& 69.2& 85.2& 57.4& 58.4& 43.6& 60.1& 47.7& 58.6\\
Hossain \etal~\shortcite{Hossain:2018}($\dagger$) &48.4& 50.7& 57.2& 55.2& 63.1& 72.6& 53.0& 51.7& 66.1& 80.9& 59.0& 57.3& 62.4& 46.6& 49.6& 58.3\\
Pavlakos \etal~\shortcite{pavlakos2018ordinal}($+$)& 48.5& 54.4& 54.4& 52.0& 59.4& 65.3& 49.9& 52.9& 65.8& 71.1& 56.6& 52.9& 60.9& 44.7& 47.8& 56.2\\
\dl{Lee \etal~\shortcite{Lee:2018}}($\dagger$)& 43.8& 51.7 & 48.8 & 53.1 & 52.2 & 74.9 & 52.7 & 44.6 & 56.9 & 74.3 & 56.7 & 66.4 & 47.5 & 68.4 & 45.6 & 55.8\\
Habibie \etal~\shortcite{Habibie:2019}& 46.1& 51.3& 46.8& 51.0& 55.9& 43.9& 48.8& 65.8& 81.6& 52.2& 59.7& 51.1& 40.8 & 54.8& 45.2& 53.4\\
Chen \etal~\shortcite{Chen:2019} &43.8& 48.6& 49.1& 49.8& 57.6& 61.5& 45.9& 48.3 &62.0& 73.4& 54.8& 50.6& 56.0& 43.4& 45.5& 52.7\\
\dl{Mehta \etal~\shortcite{Mehta:2019:Xnect}}($+$) & 50.2 &61.9 &58.3 &58.2 &68.8 &74.6 &54.1 &61.5 &76.8 &91.7 &63.4 &58.5 &48.3 &65.3 &53.2 &63.6\\
Pavllo \etal~\shortcite{Pavllo:2019}($\dagger$) & 45.2 &46.7 &43.3 &45.6 &48.1 &55.1 &44.6 &44.3 &57.3 &65.8 &47.1 &44.0 &49.0 &32.8 &33.9 &46.8\\
\midrule
\dl{Single 1-D conv, gt-skeleton($\dagger$)} &55.89 &61.67 &58.88 &62.43 &62.10 &61.36 &60.57 &64.00 &72.74 &63.38 &75.27 &63.51 &58.87 &67.69 &58.87 &63.15\\
Ours, gt-skeleton($\dagger$) &43.61 &49.39 &46.60 &50.15 &49.82 &49.08 &48.29 &51.72 &60.46 &51.10 &62.99 &51.23 &46.59 &55.41 &46.59 &50.87\\
\dl{Casual conv(real-time)($\dagger$)} &51.85 &57.63 &54.84 &58.39 &58.06 &57.32 &56.53 &59.96 &68.70 &59.34 &71.23 &59.47 &54.83 &63.65 &54.83 &59.11\\
Ours($\dagger$) &47.33 &53.11 &50.32 &53.87 &53.54 &52.80 &52.01 &55.44 &64.18 &54.82 &66.71 &54.95 &50.31 &59.13 &50.31 &54.59
\end{tabular}}
\caption{Quantitative comparison to other joint position reconstruction methods using MPJPE on Human3.6M dataset. Legend: ($\dagger$) multi-frames as an input, ($+$) extra training data (Yang \etal~\shortcite{Yang:2018} use 2D annotations from the MPII dataset. Pavlakos \etal~\shortcite{pavlakos2018ordinal} use additional data from the Leeds Sports Pose (LSP) dataset).}
\label{tab:quant_human36}
\end{table*}

\begin{table}[h]
\resizebox{0.98\columnwidth}{!}{
\begin{tabular}{l|lll|lll}
Methods& \multicolumn{3}{c}{Walk} & \multicolumn{3}{c}{Jog} \\
& S1     & S2     & S3     & S1     & S2     & S3    \\
\midrule
\dl{Pavlakos \etal~\shortcite{Pavlakos:2017} }& 22.3&15.9&\textbf{22.7}&28.9&21.9&23.8\\
\dl{Pavlakos \etal~\shortcite{pavlakos2018ordinal} ($+$)}& 18.8 &12.7& 29.2 &23.5& 15.4& 14.5\\
\dl{Lee \etal~\shortcite{Lee:2018} }& 18.6 &19.9& 30.5& 25.7 &16.8 &17.7\\
\dl{Pavllo \etal~\shortcite{Pavllo:2019} }& \textbf{13.9}& \textbf{10.2} &46.6& 20.9& \textbf{13.1} &\textbf{13.8} \\
\midrule
\dl{Ours }& 15.3& 11.7 &33.7& \textbf{19.8}& 14.5 &16.1
\end{tabular}
}
\caption{\dl{Quantitative comparison to other joint position reconstruction methods using P-MPJPE on HumanEva-I. ($+$) extra training data.}}
\label{tab:quant_humaneva}
\end{table}

\paragraph{Qualitative comparison}

\begin{figure*}[h]
	\includegraphics[width=\textwidth]{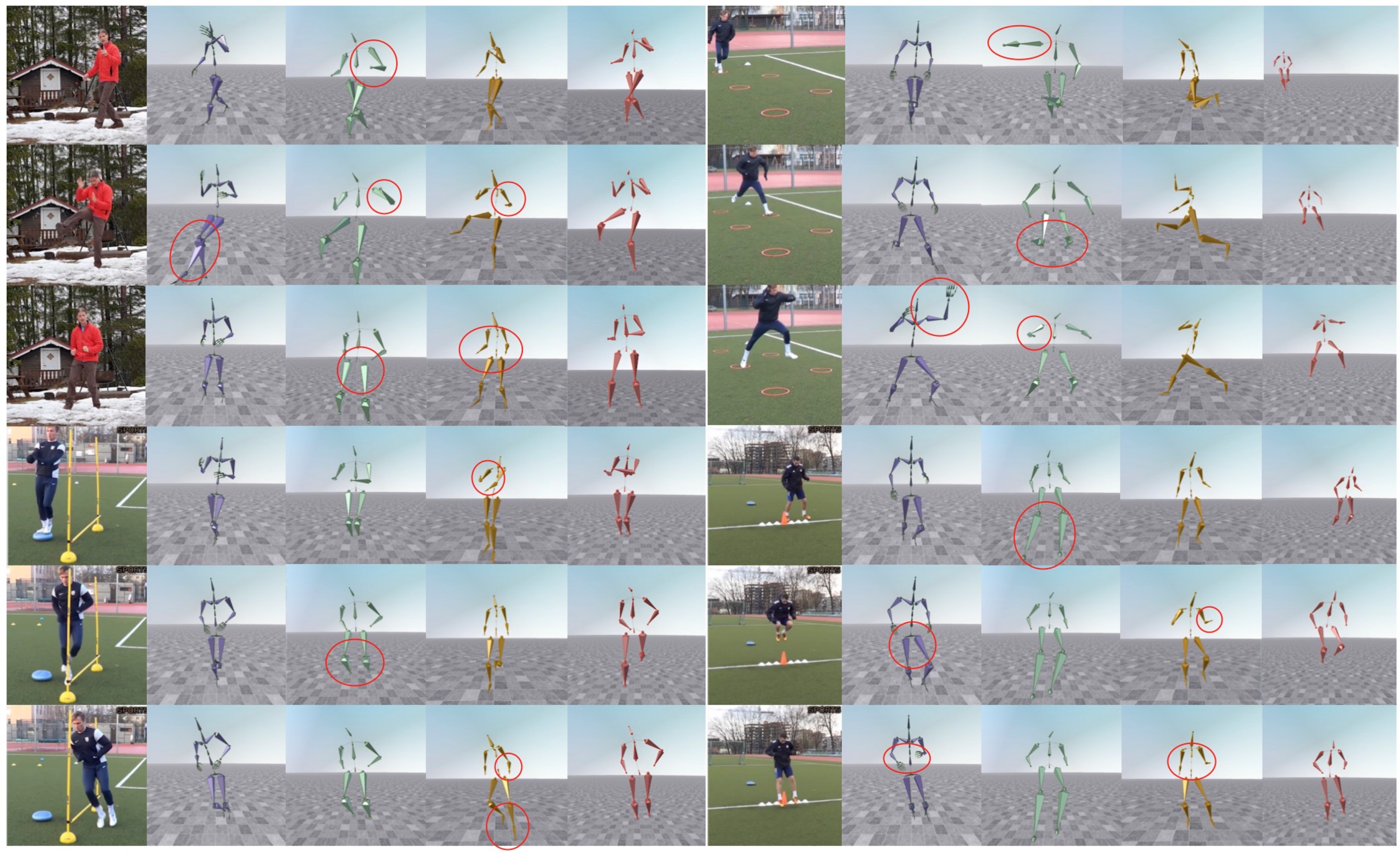}
	\resizebox{\textwidth}{!}{
	\begin{tabular}{c@{\hspace{25mm}}c@{\hspace{2mm}}c@{\hspace{2mm}}c@{\hspace{12mm}}c@{\hspace{10mm}}c@{\hspace{25mm}}c@{\hspace{2mm}}c@{\hspace{2mm}}c@{\hspace{12mm}}c@{\hspace{10mm}}}
		& \cite{Kanazawa:2018} & \cite{Pavllo:2019} & \cite{Mehta:2017} & Ours &  & \cite{Kanazawa:2018} & \cite{Pavllo:2019} &\cite{Mehta:2017} &  Ours\\
	
	\end{tabular}}
\caption{Our results on videos in the wild, compared to three other methods: (i) Kanazawa \etal~\shortcite{Kanazawa:2018} (ii) Mehta \etal~\shortcite{Mehta:2017} (iii) Pavllo \etal~\shortcite{Pavllo:2019} + Dong \etal~\shortcite{Dong:2017}. Please see the supplementary video, where it may be seen that our method recovers plausible global motions with smooth joint rotations.}
\label{fig:positions_wild}
\end{figure*}

%
We next conducted qualitative comparison with methods that aim to reconstruct joint rotations. A qualitative quantitative comparison is not enabled in that case since these different methods use a different number of joints and the rotation values might be ambiguous. The comparison is done against 3 methods: VNECT \cite{Mehta:2017}, HMR \cite{Kanazawa:2018} and Pavllo \etal~\shortcite{Pavllo:2019}, and may be seen in the supplementary video.

To compare the reconstructed motions qualitatively to Pavllo \etal~\shortcite{Pavllo:2019}, which output joint positions, the positions have to be converted to rotations. This conversion task is challenging due to ambiguities that are caused by the absence of the roll axis, temporally inconsistent skeleton size, and bone length violations that amplify the reconstruction errors and lead to abnormalities in the resulting animation.


In our comparison, we employ the rudimentary method of Dong~\etal~\shortcite{Dong:2017}, which iteratively computes the local joint angles using intrinsic rotations. 

Mehta \etal~\shortcite{Mehta:2017} use Inverse kinematics (IK) to calculates joint rotations and fit a single skeleton to given joint positions. Compared to position-to-rotation conversion using intrinsic rotations~\cite{Dong:2017}, IK yields more stable rotations suitable for animation, however, it limits the degrees of freedom to regulate the joint rotations, restricting the nature of motion.

Kanazawa~\etal~\shortcite{Kanazawa:2018} learn joint rotations in an adversarial manner in a per frame fashion. While their approach is also data driven, with no temporally consistent rotations, an animation can not be extracted directly from their per frame output. In addition, in order to extract an animation, an average skeleton should be calculated. 

The full comparison can be found in the supplementary video, and some extracted frames are depicted in Figure~\ref{fig:positions_wild}.  It can be seen that the combination of Pavllo \etal~\shortcite{Pavllo:2019} and Dong \etal~\shortcite{Dong:2017} exhibits abnormalities in the rotations resulting from the converted rotations, while the animations extracted by MotioNet do not suffer from the same issues. In addition, while Kanazawa~\etal~\shortcite{Kanazawa:2018} produce inconsistent rotations, especially next to the end-effectors (as evidenced by shaking limbs), our reconstruction yields plausible stable global motions. Due to their per-frame adversarial loss, the rotations reconstructed by their method are not temporally coherent. In order to extract a stable animation form their method, temporal smoothing and IK optimization is required, as has been done in Peng~\etal~\shortcite{peng2018sfv}.


%

Overall, these experiments show that MotioNet can recover joint positions comparably to state-of-the-art pose estimation methods, while generating more natural, temporally consistent joint rotations and plausible animations, compared to other existing methods.

\begin{figure}[h]
	\centering
	\includegraphics[width=0.90\columnwidth]{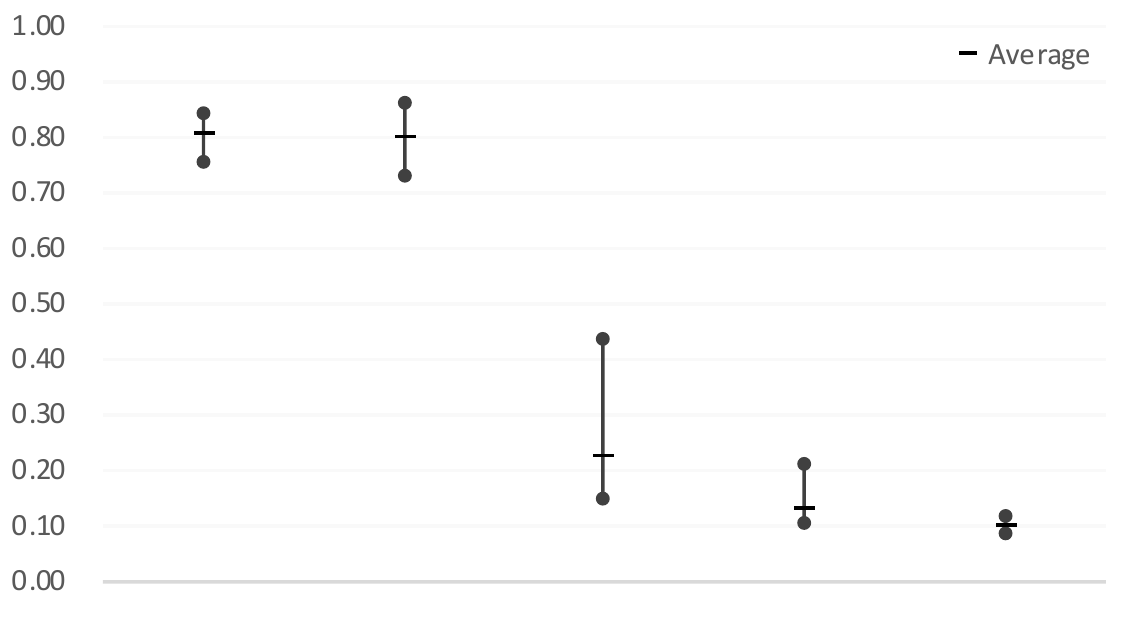}
	\begin{center}
	\resizebox{0.85\columnwidth}{!}{
	\begin{tabular}{ ccccc }
	\hspace{0.2cm}  Martinez \etal & Pavllo \etal & Mehta \etal \hspace{0.1cm} & Kanazawa \etal & Ours \hspace{1.2cm}\\
	\hspace{0.1cm} ~\shortcite{Martinez:2017} & \hspace{0.01cm} ~\shortcite{Pavllo:2019} & \hspace{0.01cm} ~\shortcite{Mehta:2017} & \hspace{0.01cm}~\shortcite{Kanazawa:2018} &  \hspace{1cm}  \\
	\end{tabular}
	}
	\end{center}
	
	\caption{Bone lengths estimation. The maximum, minimum and average bone length errors calculated for Human3.6M test subjects.}
	\label{tab:bone_length}
\end{figure}

\paragraph{Bone Lengths Estimation}
In this experiment we evaluate the accuracy of the bone lengths estimation and compare the results with other methods, including VNECT \cite{Mehta:2017}, HMR \cite{Kanazawa:2018}, Martinez \etal~\shortcite{Martinez:2017}, and Pavllo \etal~\shortcite{Pavllo:2019}. We tested the methods using Human3.6M and measured the results against the ground truth skeleton, where each bone length was calculated by the Euclidean distance between the two relevant joints. In order to avoid the scaling bias, a normalization is applied to all the skeletons during training, while at test time we de-normalize the output skeletons using the saved scaling factor and calculate the error against the original ground truth.
The final error values, which are calculated as the average of all the bone errors, are plotted in Figure~\ref{tab:bone_length}. It can be seen that our network $E_S$ outperforms previous methods in the task of bone lengths estimation.

\begin{figure}[t]
	\centering
	\includegraphics[width=0.94\columnwidth]{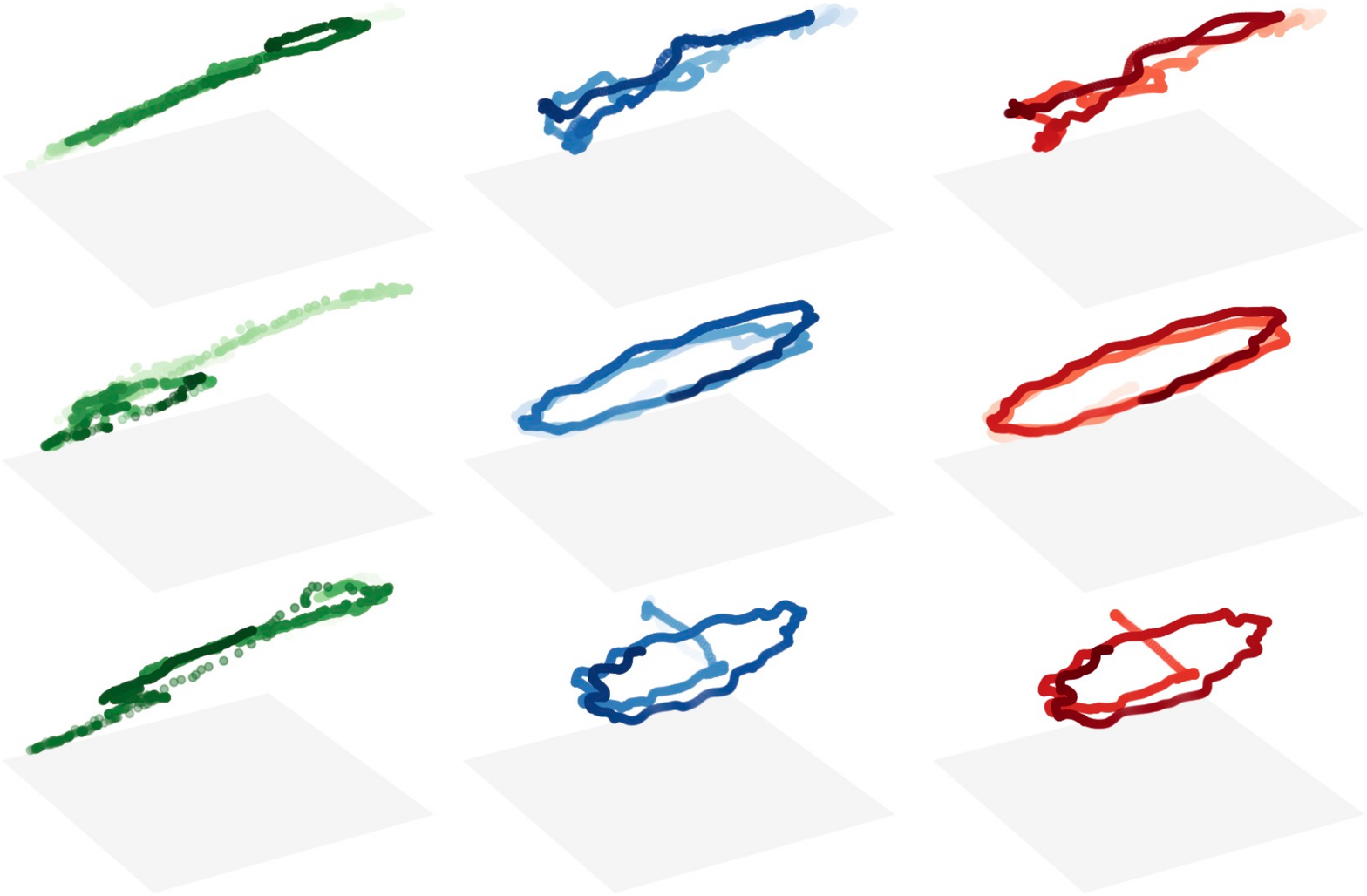}
	\begin{tabular}{ccc}
	VNECT \shortcite{Mehta:2017}  &  \hspace{1cm} Ours  \hspace{1.5cm} & GT \hspace{0.8cm}
	\end{tabular} 
	\caption{
	Comparison of global root position reconstruction accuracy. The estimated (VNECT \cite{Mehta:2017} - left column, ours - middle column) and the GT positions (right column), $(X_r^t, Y_r^t, Z_r^t)$, were extracted form 3 motions in the test data (top to bottom): directions, walking and eating.}
\label{fig:global_positions}
\end{figure}

\paragraph{Global Root Positions Reconstruction}
In this experiment we evaluate the reconstruction of the global root positions and compare it to VNECT \cite{Mehta:2017}. The evaluation was done using the test set of Human3.6M. Since the scaling factor of the reconstructed root position of VNECT is unknown, we avoid a quantitative comparison in that case. In order to qualitatively evaluate the reconstruction we visualize the positions of the root coordinate along time for some motions, where the marker becomes darker when time progresses, as shown in Figure~\ref{fig:global_positions}. The root path is visualized from another point of view (different than the camera's perspective) to demonstrate that our depth estimation is reliable (which can not be observed from the camera perspective). More qualitative results can be found in supplementary video.

\paragraph{Online Motion Reconstruction}
\dl{In this experiment we demonstrate the potential of our framework to be a core part of an online motion reconstruction system. In order to conduct such an experiment, we replaced our convolutions with casual convolutions, namely, convolution kernels that consider only past frames, while the kernels keep their original dimensions. The results in Table~\ref{tab:quant_human36} show that such a setup still enables to achieve good reconstruction. However, note that the runtime during inference, in practice, also depends on other factors, such as the 2D key-points extractor.}

\paragraph{Rotation Representation}
3D rotations may be represented with Euler angles (3D) or unit quaternions (4D). Recently, Zhou \etal~\shortcite{Zhou_2019_CVPR} proposed a new 6D representation for rotations, which might be helpful for training deep neural networks due to the inherent discontinuities that exist in other representations. Since our framework is not limited to a specific representation, we conducted an experiment that compares between these three representations. We found that the 6D representation of Zhou \etal~\shortcite{Zhou_2019_CVPR} achieves the lowest quantitative error (52.54), which is comparable to the error that is achieved by the unit quaternions (52.69), but much smaller than Euler angles (116.33). However, by qualitatively evaluating the results shown in the supplementary video, it can be seen that in practice, the results with the 6D representation, contain self-rotations and minor global shaking (note that the visualization is done without the global root position, which is irrelevant to the comparison). In contrast, in the quaternion results this phenomenon is mitigated. For the given structure of the network, it seems that the network is more challenged in the production of stable results when the output matrix is larger. Thus, we pick unit quaternions as our representation through all the experiments.

\subsection{Ablation Study}
\label{subsec:ablation}
In this section we evaluate and demonstrate the effectiveness of different components and losses in our framework.

\paragraph{Simulated Confidence Map and Augmentation}
In this experiment we compare between our results in two different setups: with and without the simulated confidence map.
In order to measure the results, we used the Human3.6M dataset, which additionally to the motion files contains video frames of the motion captured actors. 
The frames enable to simulate the full pipeline that consists of 2D position extraction (instead of 2D projection), and to evaluate the results against ground truth. We used 239 raw videos from the test data, extracted the 2D joint positions using ~\cite{Cao:2018}, evaluated the 3D output against the GT and reported the results in Table~\ref{tab:ablation}. It can be seen that although the confidence map consideration degrades the results over the clean data, it improves the results of the 2D joints extracted from videos.
A qualitative comparison can be found in the supplementary video. It can be observed that without the simulated confidence map, training over clean projected data, leads to minor shakes, which are caused by the noisy input extracted from the video frames and don't appear in the training data.
In addition, we perform an experiment without any data augmentation \dl{and one without the T-pose loss}, and report the results in Table~\ref{tab:ablation}, which demonstrate the contribution of each step.

\begin{figure}[h]
	\centering
	\includegraphics[width=0.94\columnwidth]{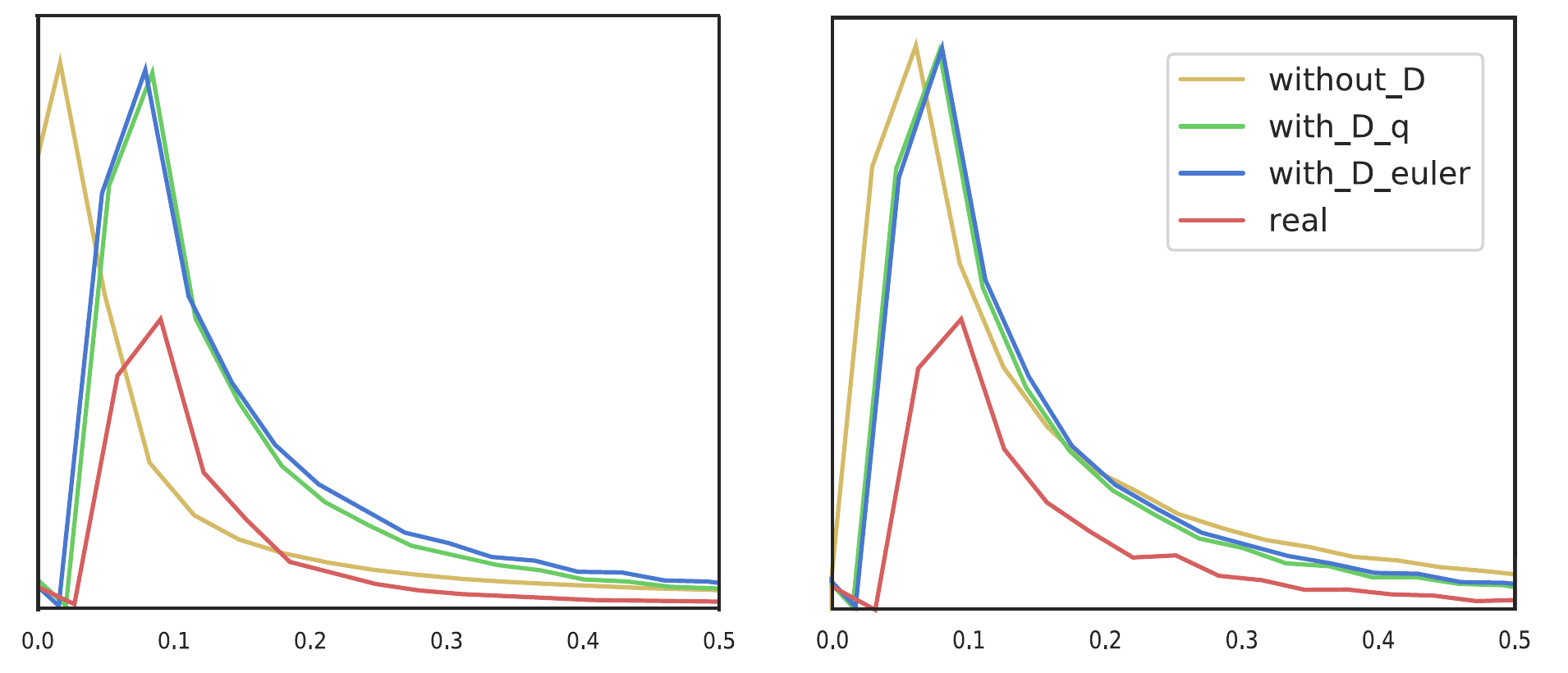}
	\begin{tabular}{cc}
	 \hspace{1.5cm}Left foot&  \hspace{2.4cm} Left Hand \hspace{3.4cm}
	\end{tabular} 
	\caption{
	Distribution of the magnitude of the joint rotational velocity of the left foot (left) and left hand (right) in our CMU test data: GT (red), outputs without a discriminator (yellow), outputs with a discriminator applied to rotations represented in unit quaternions (green) and outputs with a discriminator applied to rotations represented in Euler angle (blue).
	}
	\label{fig:rotation_distrubution}
\end{figure}

\paragraph{Rotational Velocities Discriminator}
Here we evaluate the quality of the motion reconstruction without the $\Loss_{\text{Q_{GAN}}}$ loss term, using the CMU test dataset which contains joint rotations for every sample.
A qualitative comparison can be found in the supplementary video.  Although the rotations are temporally smooth due to the existence of temporal convolution filters, it can be seen that the absence of the discriminator in our framework might yield unnatural self rotations (especially around the roll axis) and inconsistent rotation velocities. 
Figure~\ref{fig:rotation_distrubution} demonstrates the distribution of the temporal differences of the learned rotations in three different setups: without a discrimina tor, with a discriminator applied to rotations represented in unit quaternions and a discriminator applied to rotations represented in Euler angles. In all of the last experiments, the network outputs unit quaternions, while appropriate conversions are performed for the Euler angles experiment. The diagram depicts the quantity of the magnitude of the rotational velocity, per axis, in Euler angles representation. It can be seen that the distribution of the velocities become closer to the distribution of the original data with the attendance of the discriminator, for both representations, even though no direct loss is applied to the output rotations. In addition, Table~\ref{tab:ablation} shows that the quaternion-based discriminator quantitatively performed better than the Euler angle experiments, and this is our chosen representation for the rest of our experiments.

\begin{table}[h]
\resizebox{\columnwidth}{!}{
\begin{tabular}{l|ll}
\dl{Method} & Error (GT) & Error (CPN FT) \\
\hline
Baseline & 52.69 & 54.59  \\
Baseline with augmentation & 52.32 & 54.41  \\
Baseline with discriminator (quaternions) & 69.78 & 73.48  \\
Baseline with discriminator (Euler) & 70.23 & 74.14 \\
\dl{Baseline without T-pose loss} & 55.08 & 57.81  \\                              
\end{tabular}
}
\caption{\dl{Ablation study}. Various comparisons to demonstrate the effectiveness of different components and losses in our framework against the baseline. The second column depicts the error obtained by clean 2D inputs (projections of the 3D data), while the values in the third column obtained form inputs that were extracted from real video frames using fine-tuned CPN detector.}
\label{tab:ablation}
\end{table}

\begin{figure}
	\centering
	\includegraphics[width=0.94\columnwidth]{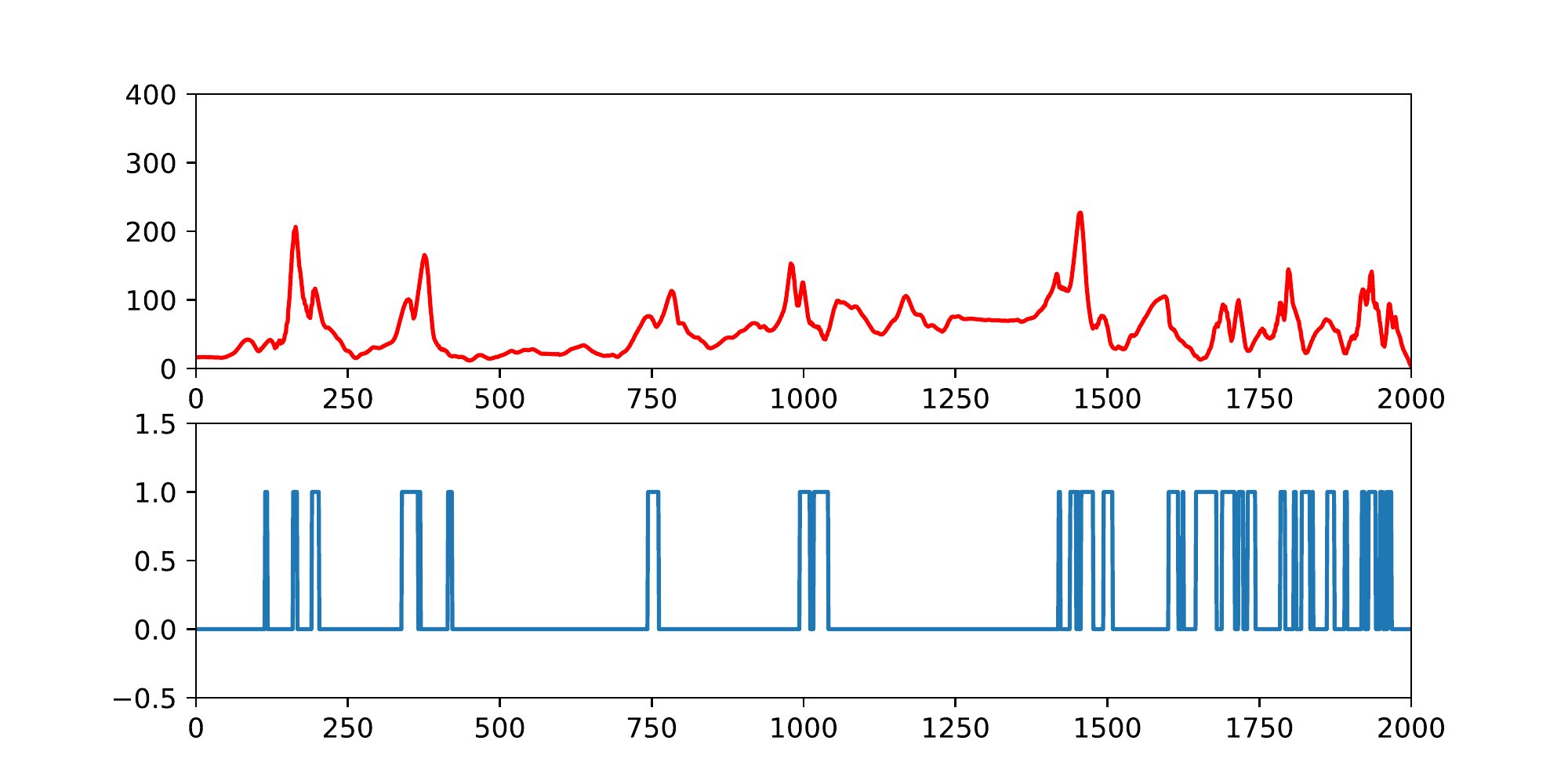}
	\caption{
	Foot contact label prediction is correlated with foot positional error. Top: foot positional error, bottom: foot contact label prediction accuracy (0 - correct, 1 - incorrect).
	}
	\label{fig:foot_error}
\end{figure}

\paragraph{Foot Contact Loss}
We performed an experiment where the network doesn't consider the foot contact prediction and foot contact loss and show its results in the supplementary video. It can be seen that when no explicit attention is given to foot contact, perceptually disturbing foot skating can be easily observed. In particular in our case, where FK is applied in a successive order from the root to the end-effectors, the accumulated error makes the unstable feet even more noticeable. 
The experiment results in a foot contact error of $64.01$ vs. error of $52.34$ with the foot contact loss, while the error is calculated by the Euclidean distance between the output and the GT foot positions only in frames where there is a contact with the ground and averaged along time. Qualitative evaluation in the supplementary video demonstrates that the prediction helps to mitigate foot sliding artifacts. In addition, as visually demonstrated in Figure~\ref{fig:foot_error} (for some arbitrary slot of motion from our test data), our experiments show that there is a high correlation between the accuracy of the predicted foot label (bottom) to the error of the foot position (top). Despite the fact that the network doesn't make use of the predicted foot contact labels, high contact prediction accuracy yields better accuracy for feet positions.

%

\section{Conclusion, limitations, and future work}
\label{sec:discussion}


We introduce a holistic approach that learns to extract a complete 3D human motion from monocular video, instead of inferring a sequence of human poses. Using standard skinning and post-processing, similar to the ones applied to data that was captured in a controlled environment, we can achieve even higher quality animations that can be used in various applications and game engines, as can be seen at the end of our supplementary video. Specifically, while previous works perform pose estimation, reconstructing only 3D joint positions, our deep neural network, MotioNet, is trained to recover a \emph{single} skeleton with a sequence of 3D joint \emph{rotations}, and, consequently, the joint positions, as well. In general, joint positions do not fully capture human motions. Human poses defined by joint positions alone may not satisfy the fixed bone length constraint. Even with the use of IK, ambiguities and temporal discontinuities can still occur. In contrast, a fixed skeleton with a temporal sequence of designated joint rotations completely characterizes the motion; it is a native representation for animation synthesis, as it often composes the subspace of the character rig space.

A key advantage of our approach is that it does not require IK: both joint positions and joint rotations are learned from training data of real human motions, rather than solved for. This leads to a more natural human motion reconstruction, via the FK layer of MotioNet. Furthermore, by inferring a single skeleton, MotioNet is guaranteed to keep the bone lengths fixed along the entire motion sequence. This provides a prior that facilitates motion reconstruction in the presence of significant occlusion in the input video. Another mechanism we employ to address the occlusion challenge is to inject random noise to the training input sequences and augment them with confidence values whose distribution mimics that of confidence values extracted by 2D pose estimation methods from videos in the wild. The network is thus trained to cope with noise and occlusions, akin to denoising autoencoders.
Finally, since our system is trained in the space of motions, the inherent smoothness of human motions is learned from the data, achieving temporal coherence naturally. 


By design, MotioNet only tracks the motion of a single human character. The handling of multi-character motions, especially when the characters undergo close interactions, is clearly challenging, and we leave it for future work. Also, our current method may not accurately recover the global positioning of the skeletons, e.g., when there are significant camera movements. 
Another limitation is that our network does not account for any physical constraints of the human character or of the interactions between the character and the environment --- the motion reconstruction is purely based on analyzing and learning from visual cues, which could be billed as an advantage as well. As a result, for example, there is no notion of contact constraints in the system, thus the resulting motion may suffer from foot sliding. This could potentially be solved by predicting the foot contact conditions and using a loss function term that evaluates the positional constraints as done by Lee~\etal~\shortcite{Lee:2018:ICA:3272127.3275071}. 
%

%

In other possible future works, we are interested in learning the subspace of human motion from the videos, so that animators can easily produce physically-plausible animations with little effort. We are also interested in learning the differences between motion styles from a large amount of video data that is currently publicly available. A dataset with motion style annotations will be useful for creating a generative model, where novel motion styles can be sampled. 

\label{sec:acknowledgments}
\begin{acks}
We thank the anonymous reviewers for their constructive comments. This work was supported in part by the National Key R\&D Program of China (2018YFB1403900, 2019YFF0302902), the Israel Science Foundation (grant no.~2366/16), and by the European Union's Horizon 2020 Research and Innovation Programme under Grant Agreement  No 739578 and the Government of the Republic of Cyprus through the Directorate General for European Programmes, Coordination and Development.
\end{acks}

\bibliographystyle{ACM-Reference-Format}
\bibliography{References_new} 

\appendix
\section{Network Architecture}
\label{appendix}
\dl{In this part we describe the layers of the components in our architecture ($E_S$, $E_Q$ and $D$) and list them in Table~\ref{tab:architecture}.}

\dl{In both $E_S$ and $E_Q$ there are three main parts: channel-expansion, information-fusion and channel-shrink. In the expansion and shrink parts, we use convolutional blocks, where the convolutions have kernel size 1 and stride 1. In the information-fusion part, for $E_Q$, we apply three parallel convolutional blocks, where each has a different kernel size, and then use adaptive average pooling to ensure that the temporal length of the output sequence is equal to the input one. For $E_s$, we use sequential convolution blocks and then collapse all the temporal information into a single vector with adaptive max pooling.}
 
 Our discriminator $D$ is a linear component (similarly to Kanazawa~\etal~ \shortcite{Kanazawa:2018}), with an output value between 0 and 1, containing two convolution layers and one fully connected layer. 
 
\dl{In each convolution block, we use BatchNorm, LReLU, Dropout. The kernel width and the stride are denoted by $k$ and $s$, respectively, and the number of input and output channels is given in the rightmost column.}
\begin{table}[h]
\resizebox{\columnwidth}{!}{
\begin{tabular}{| l | c | l | c | c | c |}
  \hline
  \toprule
      Name & & Layers & $k$ & $s$ & in/out \\
  \hline
  \midrule 
      $E_Q$  & &\texttt{Conv + BatchNorm + LReLU + Dropout}  & $1$ & 1 & $34/2048$\\
      \cline{2-5}
       & $\rightarrow$ &\texttt{Conv + BatchNorm + LReLU + Dropout + Adap AP} & $5$ & 3 & $2048/2048$\\
      & $\rightarrow$ &\texttt{Conv + BatchNorm + LReLU + Dropout + Adap AP} & $3$ & 1 & $2048/2048$\\
      & $\rightarrow$ &\texttt{Conv + BatchNorm + LReLU + Dropout + Adap AP} & $1$ & 1 & $2048/2048$\\
      \cline{2-5}
      & &\texttt{Conv} & $1$ & 1 & $1024/69$\\
  \hline
  \midrule 
      $E_S$ & & \texttt{Conv + BatchNorm + LReLU + Dropout}  & $1$ & 1 & $34/2048$\\
      \cline{2-5}
       & &\texttt{Conv + BatchNorm + LReLU + Dropout} & $5$ & 1 & $2048/2048$\\
      & $\downarrow$ &\texttt{Conv + BatchNorm + LReLU + Dropout} & $3$ & 1 & $2048/2048$\\
      & &\texttt{Conv + BatchNorm + LReLU + Dropout} & $1$ & 1 & $2048/2048$\\
      & &\texttt{Adaptive MP} & - & - & -\\
      \cline{2-5}
      & &\texttt{Conv} & $1$ & 1 & $2048/9$\\
  \hline
	\midrule 
      $D$ & & \texttt{Conv}  & $1$ & 1 & $4/16$\\
       & $\downarrow$&\texttt{Conv} & $1$ & 1 & $16/16$\\
      & &\texttt{Linear} & $-$ & - & $16/1$\\
  \hline
  \bottomrule
\end{tabular}

} 
\caption{\dl{MotioNet structure. $\rightarrow$ denotes parallel convolutions while $\downarrow$ denotes sequential ones.}}
\label{tab:architecture}
\end{table}


\end{document}